\documentclass[sigconf]{acmart}

\copyrightyear{2024}
\acmYear{2024}
\setcopyright{acmcopyright}\acmConference[ICCAD '24]{IEEE/ACM International Conference on Computer-Aided Design}{October 27-October 31, 2024}{New Jersey, USA}
\acmBooktitle{IEEE/ACM International Conference on Computer-Aided Design (ICCAD '24), 2024}
\acmPrice{15.00}
\acmDOI{10.1145/3508352.3549478}
\acmISBN{978-1-4503-9217-4/22/10}

\usepackage{amsmath,amssymb,amsfonts}
\usepackage{graphicx}
\usepackage{algorithm}
\usepackage{algpseudocode}
\usepackage{threeparttable}
\usepackage{graphics}
\usepackage{epsfig}
\usepackage{textcomp}
\usepackage{xcolor}
\usepackage{amssymb}
\usepackage{booktabs}
\usepackage{multirow}
\usepackage{makecell}

\usepackage{hyperref}
\usepackage{url}
\usepackage{mathtools}

\usepackage{natbib}

\AtBeginDocument{%
  \providecommand\BibTeX{{%
    \normalfont B\kern-0.5em{\scshape i\kern-0.25em b}\kern-0.8em\TeX}}}

\usepackage{enumitem}
\setlist[itemize]{leftmargin=0.5cm}

\title{Co-Designing Binarized Transformer and Hardware Accelerator for Efficient End-to-End Edge Deployment} 

\author{Yuhao Ji$^{*\thanks{*~The first two authors contribute equally to this work.},1,2}$, Chao Fang$^{*,1}$, Shaobo Ma$^1$, Haikuo Shao$^1$, and Zhongfeng Wang$^{1,3}$}

\affiliation{
\institution{
$^1$School of Electronic Science and Engineering, Nanjing University, Nanjing, China \\
$^2$Department of Computer Science and Engineering, Chinese University of Hong Kong, Hong Kong, China  \\
$^3$School of Integrated Circuits, Sun Yat-sen University, Shenzhen, China}
\country{}
}

\email{
arthurxxzh@gmail.com, {fantasysee, 201180102, hkshao}@smail.nju.edu.cn, zfwang@nju.edu.cn
}

\def \authors{Yuhao Ji, Chao Fang, Shaobo Ma, Haikuo Shao, Zhongfeng Wang}
\renewcommand{\shortauthors}{Yuhao Ji, Chao Fang, Shaobo Ma, Haikuo Shao, Zhongfeng Wang}

\usepackage{pifont}
\usepackage{xcolor} 
\usepackage{graphicx}
\usepackage{hyperref}
\usepackage{cleveref}
\Crefformat{figure}{Fig.~#2#1#3}                           
\Crefname{subfigure}{Fig.}{Figs.}
\Crefname{figure}{Fig.}{Figs.}
\Crefformat{table}{TABLE~#2#1#3}                           
\usepackage{amsmath}
\usepackage{threeparttable}
\usepackage{subfig}

\newcommand{\minisection}[1]{\vspace{.1in}\noindent{\textbf{#1}}}

\newcommand{\Rone}[1]{\textcolor{black}{#1}}

\newcommand{\fc}[1]{\textcolor{black}{#1}} 
\newcommand{\shk}[1]{\textcolor{black}{#1}}
\newcommand{\wx}[1]{\textcolor{black}{#1}}
\newcommand{\update}[1]{\textcolor{black}{#1}}

\begin{document}

\begin{abstract}

\fc{Transformer models have revolutionized AI tasks, but their large size hinders real-world deployment on resource-constrained and latency-critical edge devices.}
\fc{While binarized Transformers offer a promising solution by significantly reducing model size, existing approaches suffer from algorithm-hardware mismatches with limited co-design exploration, leading to suboptimal performance on edge devices.}
\fc{Hence, we propose a co-design method for efficient end-to-end edge deployment of Transformers from three aspects: algorithm, hardware, and joint optimization.}
\fc{First, we propose BMT, a novel hardware-friendly binarized Transformer with optimized quantization methods and components, and we further enhance its model accuracy by leveraging the weighted ternary weight splitting training technique.}
\fc{Second, we develop a streaming processor mixed binarized Transformer accelerator, namely BAT, which is equipped with specialized units and scheduling pipelines for efficient inference of binarized Transformers.}
\Rone{Finally, we co-optimize the algorithm and hardware through a design space exploration approach to achieve a global trade-off between accuracy, latency, and robustness for real-world deployments.}
\fc{Experimental results show our co-design achieves up to 2.14$\sim$49.37$\times$ throughput gains and 3.72$\sim$88.53$\times$ better energy efficiency over state-of-the-art Transformer accelerators, enabling efficient end-to-end edge deployment.}
\end{abstract}

\maketitle


\section{Introduction}\label{sec:intro}

\fc{Transformer-based neural networks have experienced a remarkable surge in recent years, from BERT \cite{BERT}, to ViT \cite{ViT, UCViT}, further to the large language models (LLMs) \cite{GPT3, OPT}, demonstrating exceptional performance across a diverse range of tasks.}
\fc{However, the dramatic increase in model size and computational complexity has imposed significant constraints, particularly in latency-critical and resource-constrained end-to-end edge scenarios.}

\fc{Quantization \cite{QBERT, zafrir2019q8bert, TernaryBERT, I-BERT, I-ViT} emerges as a promising solution to alleviate the challenge of Transformer edge deployment, which reduces model size and computational complexity by lowering the bit width of weights or activations.}
Among them, \textit{binarized Transformer} \cite{BinaryBERT, BiBERT, BEBERT, BiT} stands out, which reduces the bit width of weights to 1-bit and transforms computations to bit-wise operations, minimizing both parameter storage and computational complexity. 
\fc{However, challenges remain on both algorithm and hardware 
\Rone{levels}
for efficient end-to-end edge deployment of binarized Transformers.}

\Rone{On the algorithm level, existing binarized Transformers struggle for efficient edge deployment due to}
\fc{hardware-unaware model architectures, 
\Rone{theoretical-practical performance gap,}
and neglected model robustness.}
\fc{1) Existing binarized Transformers often lack consideration for hardware implementation, 
\Rone{posing challenges to the deployment on edge devices with limited resources.}
For instance, the use of complex activation functions like GELU in BinaryBERT \cite{BinaryBERT} demands significant hardware resources and sophisticated design \cite{xiaowu2023nonlinear}, increasing implementation costs.  
}
\fc{2) Binarized Transformers often prioritize theoretical compression ratios, neglecting the impact on hardware performance. 
\Rone{For instance, existing works~\cite{BinaryBERT, BiT, BEBERT} only evaluate the theoretical reduction in model size and FLOPs of their binarized Transformers without validating the actual speedup performance.}
Consequently, these models may not translate theoretical compression benefits into real-world performance improvements on edge devices.}
\fc{3) 
\Rone{While accuracy has improved, the robustness of binarized Transformers remains hardly explored, which is crucial for ensuring reliable performance in diverse and challenging environments, especially at the edge where data quality may be various and unpredictable.}
}

\begin{table*}[t]
\centering
\caption{Comparative Analysis of the Related Works and Ours}
\label{tab:comparison}
\resizebox{0.92\textwidth}{!}{%
\begin{threeparttable}
\begin{tabular}{c|c|c|c|c|c|c}
\hline
\textbf{Work} & \textbf{Alg Tech.} & \textbf{HW Bit-width} & \textbf{HW Implementation} & \textbf{HW Arch.} & \textbf{End to end} & \textbf{Tune} \\ \hline
I-BERT \cite{I-BERT} & \textcolor{green}{\ding{51}} Quantization (W8A8$^\dagger$) & \textcolor{green}{\ding{51}} FIX8 & \textcolor{red}{\ding{55}} Tesla T4 GPU & N/A & \textcolor{green}{\ding{51}} Yes & Alg Tuning \\ \hline
I-ViT \cite{I-ViT} & \textcolor{green}{\ding{51}} Quantization (W8A8$^\dagger$) & \textcolor{green}{\ding{51}} FIX8 & \textcolor{red}{\ding{55}} RTX 2080 Ti GPU & N/A & \textcolor{green}{\ding{51}} Yes & Alg Tuning \\ \hline
BinaryBERT \cite{BinaryBERT} & \textcolor{green}{\ding{51}} Binarization (W1A8/4$^\dagger$) & \textcolor{red}{\ding{55}} FP32 & \textcolor{red}{\ding{55}} GPU & N/A & \textcolor{green}{\ding{51}} Yes & Alg Tuning \\ \hline
OPTIMUS \cite{OPTIMUS} & \textcolor{red}{\ding{55}} N/A & FIX16 & \textcolor{green}{\ding{51}} ASIC 28nm & Processor & \textcolor{red}{\ding{55}} No & \textcolor{red}{\ding{55}} N/A \\ \hline
$A^3$ \cite{A3} & \textcolor{red}{\ding{55}} N/A & \textcolor{green}{\ding{51}} FIX8 & \textcolor{green}{\ding{51}} ASIC 40nm & Streaming & \textcolor{red}{\ding{55}} No & \textcolor{red}{\ding{55}} N/A \\ \hline
FTRANS \cite{FTRANS} & \textcolor{green}{\ding{51}} BCM-based Prune & FIX16 & \textcolor{green}{\ding{51}} FPGA VCU118 & Streaming & \textcolor{green}{\ding{51}} Yes & HW Tuning \\ \hline
Lu et al. \cite{Lu} & \textcolor{red}{\ding{55}} N/A & \textcolor{green}{\ding{51}} FIX8 & \textcolor{green}{\ding{51}} FPGA VU13P & Processor & \textcolor{red}{\ding{55}} No & \textcolor{red}{\ding{55}} N/A \\ \hline
EFA-Trans \cite{EFA-Trans} & \textcolor{green}{\ding{51}} Bank-balanced Prune & \textcolor{green}{\ding{51}} FIX8 & \textcolor{green}{\ding{51}} FPGA ZCU102 & Processor & \textcolor{red}{\ding{55}} No & \textcolor{green}{\ding{51}} Alg\&HW Co-tuning \\ \hline
FQ-BERT \cite{FQ-BERT} & \textcolor{green}{\ding{51}} Quantization (W4A8$^\dagger$) & \textcolor{green}{\ding{51}} FIX8/4 & \textcolor{green}{\ding{51}} FPGA ZCU102 & Processor & \textcolor{green}{\ding{51}} Yes & \textcolor{red}{\ding{55}} N/A \\ \hline
ViA \cite{ViA} & \textcolor{red}{\ding{55}} N/A & \textcolor{red}{\ding{55}} FP16 & \textcolor{green}{\ding{51}} FPGA Alveo U50 & Streaming & \textcolor{green}{\ding{51}} Yes & HW Tuning \\ \hline
Fan et al. \cite{butterfly} & \textcolor{green}{\ding{51}} Butterfly-sparsity Prune & \textcolor{red}{\ding{55}} FP16 & \textcolor{green}{\ding{51}} FPGA Zynq 7045 & \textcolor{green}{\ding{51}} Mixed & \textcolor{green}{\ding{51}} Yes & \textcolor{green}{\ding{51}} Alg\&HW Co-tuning \\ \hline
\textbf{Ours} & \textbf{\textcolor{green}{\ding{51}} Binarization (W1A8/4/2/1$^\dagger$)} & \textbf{FP16-FIX8/4/2/1 Mixed} & \textbf{\textcolor{green}{\ding{51}} FPGA ZCU102} & \textbf{\textcolor{green}{\ding{51}} Mixed} & \textcolor{green}{\ding{51}} Yes & \textbf{\textcolor{green}{\ding{51}} Alg\&HW Co-tuning} \\
\hline
\end{tabular}%
\begin{tablenotes}
    \item[$\dagger$] A network with weights quantized to $b_w$ bits and activations quantized to $b_a$ bits is denoted as $Wb_wAb_a$.
\end{tablenotes}
\end{threeparttable}
}
\vspace{-0.4cm}
\end{table*}

\fc{On the hardware level, current accelerator designs \cite{OPTIMUS, STA, FTRANS, Lu, EFA-Trans, A3, ViA, butterfly, FQ-BERT, haikuo2024efficientViT} fall short in fully enabling efficient end-to-end edge deployment of binarized Transformers due to limitations in constrained computational support, architectural inefficiencies, and suboptimal hardware configurations.}
\Rone{1) Existing accelerators primarily focus on high-bit quantized models, neglecting the unique requirements of binarized Transformers, resulting in a lack of support for binarized acceleration. 
Furthermore, the crucial hardware implementation of the quantization process itself remains unexplored, preventing an end-to-end acceleration.}
\fc{2) Existing accelerator architectures fall into two categories: streaming-like \cite{quang2022streaming-access, Minsik2024Layerwise-tcas1, Duy2021Layer-Specific, A3, FTRANS, ViA} and processor-like \cite{jiantao2016Embedded-isfpga, zhang2019compression-electronics, Lu, OPTIMUS, EFA-Trans, FQ-BERT, STA, huang2024precision, tu2020evolver}, but neither offers an ideal solution for edge deployment with binarized Transformers. 
Streaming-like architectures, optimized for efficiency with large batches, face underutilization issues when dealing with the smaller batch sizes typically encountered at the edge.
Conversely, processor-like architectures, designed for versatility, introduce overhead with complex datapaths and scheduling for intricate Transformer operations.}
\Rone{3) Additionally, the hardware configurations of existing accelerators often rely on empirical values rather than a systematic exploration, which contributes to the suboptimal performance observed in practice. Consequently, the resulting accelerators fail to operate at their full potential\fc{, hindering efficient edge deployment.}}
\fc{To facilitate efficient end-to-end edge deployment, we propose a co-design approach within binarized Transformers and hardware accelerator that addresses the above challenges from three aspects.}
\fc{From algorithm aspects, we introduce BMT, a novel binarized Transformer model that incorporates hardware-aware quantization methods and computational components, specifically designed for efficient execution on edge devices. Additionally, we develop weighted ternary weight splitting (WTWS) to optimize the training process of BMT.}
\fc{From hardware aspects,} to fully exploit the 
\Rone{practical}
acceleration of 
\Rone{binarized Transformer,} 
we design BAT, a novel streaming-processor-mixed binarized Transformer accelerator, equipped with highly optimized computational units and scheduling pipelines to enable an efficient end-to-end inference for binarized Transformers.
\fc{From joint optimization aspects, we further push the performance boundaries by co-optimizing 
\Rone{the binarized Transformers}
and BAT through a design space exploration (DSE) approach, allowing us to find a global trade-off between accuracy, latency, and robustness for real-world deployments.}

The contributions of this work are summarized as follows:
\begin{itemize}
    \item A hardware-friendly binarized Transformer, BMT, which yields a substantial compression ratio compared to the full precision baseline \fc{with comparable accuracy}, exhibiting significant potential for edge deployment. (\Cref{sec:algo})
    \item A streaming-processor-mixed binarized Transformer accelerator, BAT, enabling high efficiency and low power end-to-end acceleration of binarized Transformers. (\Cref{sec:accel})
    \item Dataflow optimizations
    \Rone{adaptable for}
    edge scenarios 
    \Rone{to}
    improve the hardware efficiency, and a DSE approach to jointly optimize both algorithmic and hardware parameters, striking a fine balance between accuracy, robustness, and latency. (\Cref{sec:opt})
    \item \Rone{Comprehensive experiments showing our co-design achieves up to 2.14$\sim$49.37$\times$ and 3.72$\sim$88.53$\times$ improvement on throughput and energy efficiency over SOTA Transformer accelerators and improves energy efficiency by 213.82$\times$ and 174.01$\times$ compared to the CPU and GPU implementations. (\Cref{sec:eval})}
\end{itemize}


\Rone{\Cref{tab:comparison} summarizes the comparison between the related works and our work, focusing on these critical features for \textbf{edge deployment}: algorithm-level technique (Alg Tech.), operating arithmetic precision on hardware (HW Bit-width), type of hardware platform (HW Implementation) and its architecture (HW Arch.), whether the work implements the end-to-end acceleration (End-to-end), whether algorithmic or hardware parameter configuration is tuned (Tune). The comparison reveals that \textit{our work presents significant benefits and potential over prior arts for efficient edge deployment}.}

\section{Background}\label{sec:bkg}

\subsection{Transformer-based Neural Networks}

\begin{figure}[ht]
    \centering
    \includegraphics[width=0.48\textwidth]{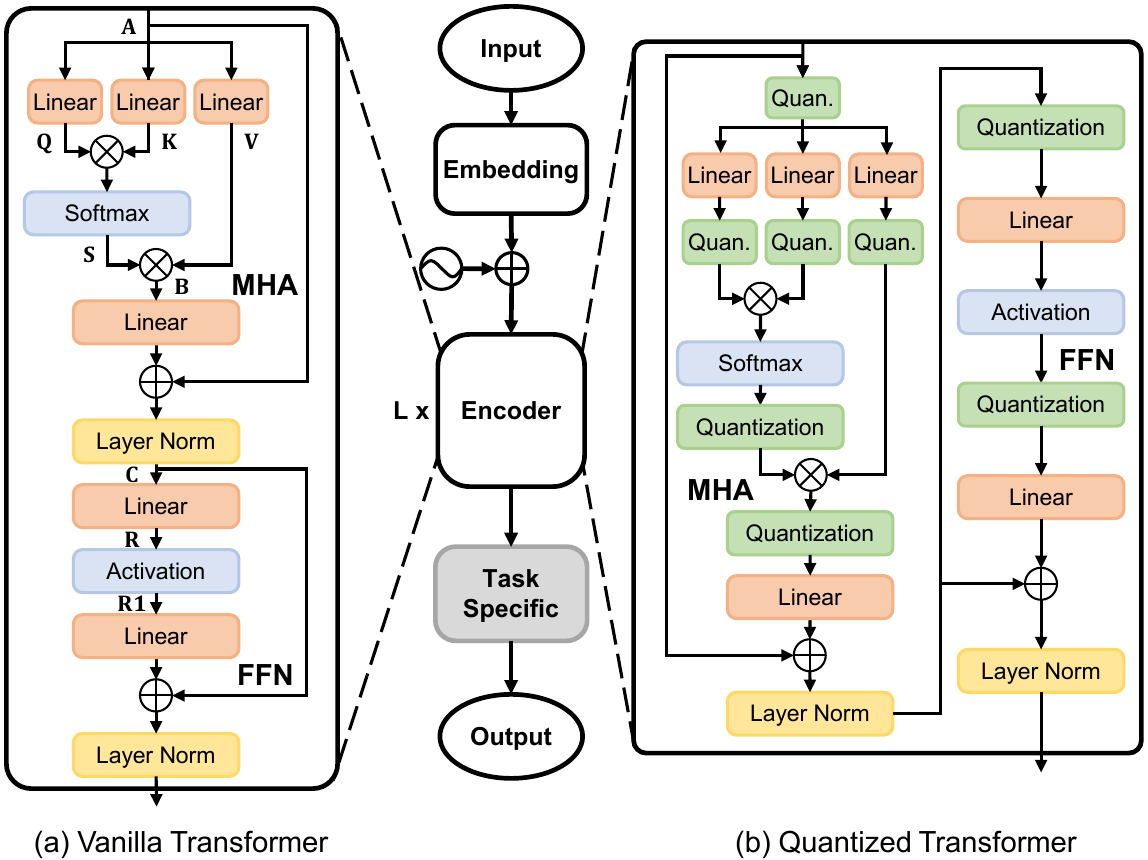}
    \caption{Overview of encoder-based Transformer structure: (a) vanilla Transformer and (b) quantized Transformer.}
    \label{fig:Transformer_Arch}
    \vspace{-0.45cm}
\end{figure}

\Cref{fig:Transformer_Arch} illustrates the architecture of the encoder-based Transformer network, where each encoder consists of a multi-head attention module (MHA) and a feed-forward network (FFN). 
Residual addition and layer normalization (LN) are used before and after FFN. Within each MHA, the input (\textbf{A}) is projected to query (\textbf{Q}), key (\textbf{K}) and value (\textbf{V}) matrices through three different linear layers (\textbf{AW}). 
The query matrix is then multiplied with $\textbf{K}^T$, and the scaled product (\textbf{Q}$\textbf{K}^T$) is processed by a softmax module to produce the score matrix (\textbf{S}). 
This matrix \textbf{S} is then multiplied with \textbf{V} and the product (\textbf{S}\textbf{V}) is passed through an additional linear layer (\textbf{BW}) to generate the final output of the MHA. 
In FFN, the input is first projected into an intermediate matrix (\textbf{R}) through a linear layer (\textbf{CW}), followed by the activation function. 
The activated result (\textbf{R1}) is then subjected to another linear transformation (\textbf{R1W}) to obtain the output of FFN. 
Compared to the vanilla Transformer, the quantized Transformer incorporates quantization processes, which quantizes weights and activations to the designed bit-width, \Rone{and matrix multiplications are transformed to quantized matrix multiplications (QMMs).} 

\subsection{Binarization and Its Unique Demands} \label{subsec:quan}


\minisection{Binarization}, which reduces the bit width to 1-bit, has been studied for a long time \cite{XNOR-Net, BWN, Bi-Real, BinaryBERT, BiT, BiBERT, BEBERT, OneBit}. A notable work, Binary-Weight-Network (BWN) \cite{BWN}, binarizes full precision weight $\boldsymbol{W_R}$ with a scaling parameter $\alpha$ according to the following equation:
\begin{equation}\label{eq:quantization}
    BWN(\boldsymbol{W_R}) = \alpha \boldsymbol{W_B},
\end{equation}
where $\boldsymbol{W_B} = Sign(\boldsymbol{W_R})$, $\alpha = \frac{1}{N} \Vert \boldsymbol{W_R} \Vert_1$, $N$ is the total number of elements of $\boldsymbol{W_R}$.

\begin{figure}[ht]
    \centering
    \includegraphics[width=0.44\textwidth]{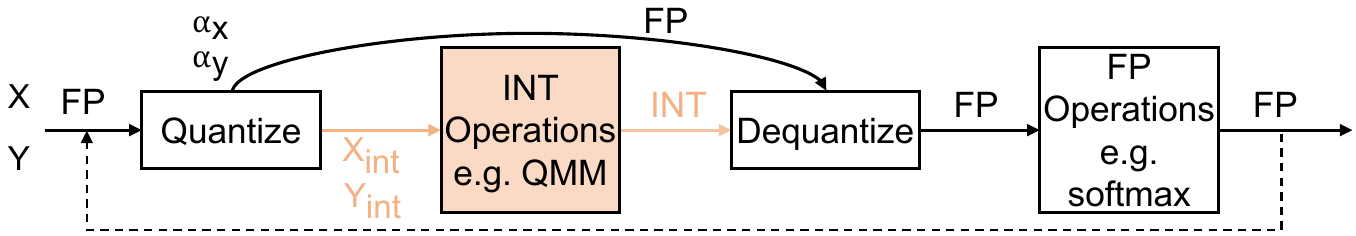}
    \caption{Quantization-dequantization computational flow.}
    \label{fig:quan-dequan}
    \vspace{-0.3cm}
\end{figure}

\fc{Binarized Transformers, while offering significant computational efficiency benefits, introduce unique demands that necessitate careful co-design of algorithms and hardware.}
\fc{This partially stems from their reliance on a quantization-dequantization scheme as shown in \Cref{fig:quan-dequan}, which enables efficient hardware acceleration by converting full-precision inputs to lower bit widths but requires dequantization for high-precision tasks.}
\fc{Binarized Transformers further complicate matters by requiring support for two distinct QMM types:}
\textit{$activation \times weight$} (linear layer) and \textit{$activation \times activation$} ($\otimes$). 
For a binarized Transformer in $W1AN$, two multiplications are involved: $N$-bit $\times$ $1$-bit and $N$-bit $\times$ $N$-bit, which is consistent with the aforementioned two QMM types. 
\update{In addition, the activation quantization may incorporate signed and unsigned quantization. }
\vspace{-0.2cm}

\section{Algorithm Optimization}\label{sec:algo}


\subsection{Hardware-friendly Model Design} \label{subsec:hw-friendly}


\fc{Limited resources in edge devices require hardware-aware design for binarized models. Existing models~\cite{BinaryBERT, BiBERT, BiT, BEBERT} lack such considerations, hindering efficient deployment. Our binarized Transformer, namely BMT, addresses this with hardware-optimized techniques.}

\minisection{Elastic Activation Quantization.} 
\fc{Existing activation quantization methods often incur high hardware resource and latency overhead due to extra scaling factor computations \cite{tim2022llmint8} during implementation on dedicated accelerators.}
\fc{To overcome this issue, we \fc{incorporate} hardware-friendly elastic quantization \cite{BiT} method for activations when designing our BMT, making it exhibit outstanding performance in the extremely low-bit scenario.} The equation can be formulated as follows:
\begin{equation}
\begin{aligned}
\boldsymbol{X_{INT}} = \alpha \lfloor clip(\frac{\boldsymbol{X_R}+\beta}{\alpha}, Q_n, Q_p) \rceil,
\end{aligned}
\end{equation}
where $\lfloor \cdot \rceil$ represents the rounding function. $\boldsymbol{X_R}$ is a full precision tensor and $\alpha$ is the full precision scaling factor. 
$Q_n$/$Q_p$ is the min/max value of the quantization range. 

\fc{BMT benefits from several aspects by using hardware-friendly quantization.}
\fc{First, both coefficients $\alpha$ and bias $\beta$ of BMT are learnable parameters, but fixed before deployment, reducing computational burden during inference.}
\fc{Second, the pre-computed reciprocal of $\alpha$ and the standardized values of $Q_n$ and $Q_p$ in BMT further streamline online inference by eliminating complex operations and simplifying the clip function implementation.}

\minisection{Non-linear Activation Function.} 
\fc{Prioritizing hardware efficiency, BMT employs the simple ReLU activation function instead of the computationally expensive GELU function commonly used in binarized Transformers. 
ReLU requires minimal hardware resources, as it can be implemented with a single multiplexer, making it ideal for resource-constrained edge devices.}


\begin{figure}[ht]
    \centering
    \includegraphics[width=0.48\textwidth]{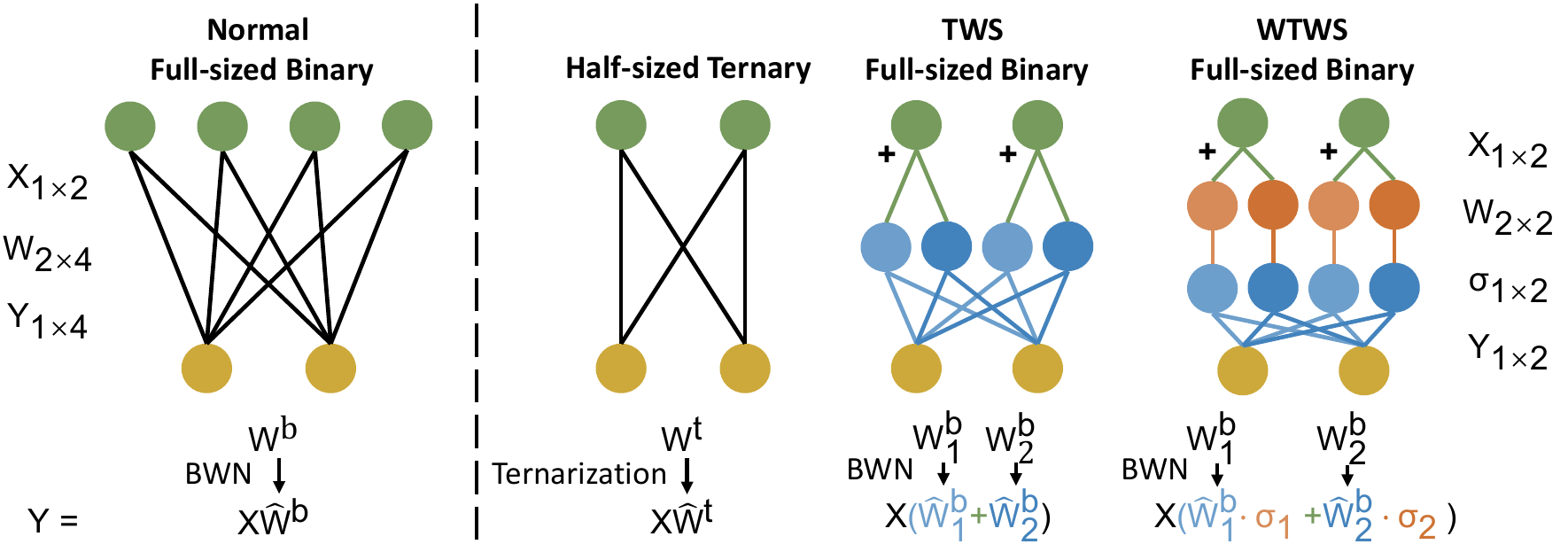}
    \caption{The illustration for WTWS.}
    \label{fig:wtws}
    \vspace{-0.4cm}
\end{figure}

\begin{figure*}[ht]
    \centering
    \includegraphics[width=0.9\textwidth]{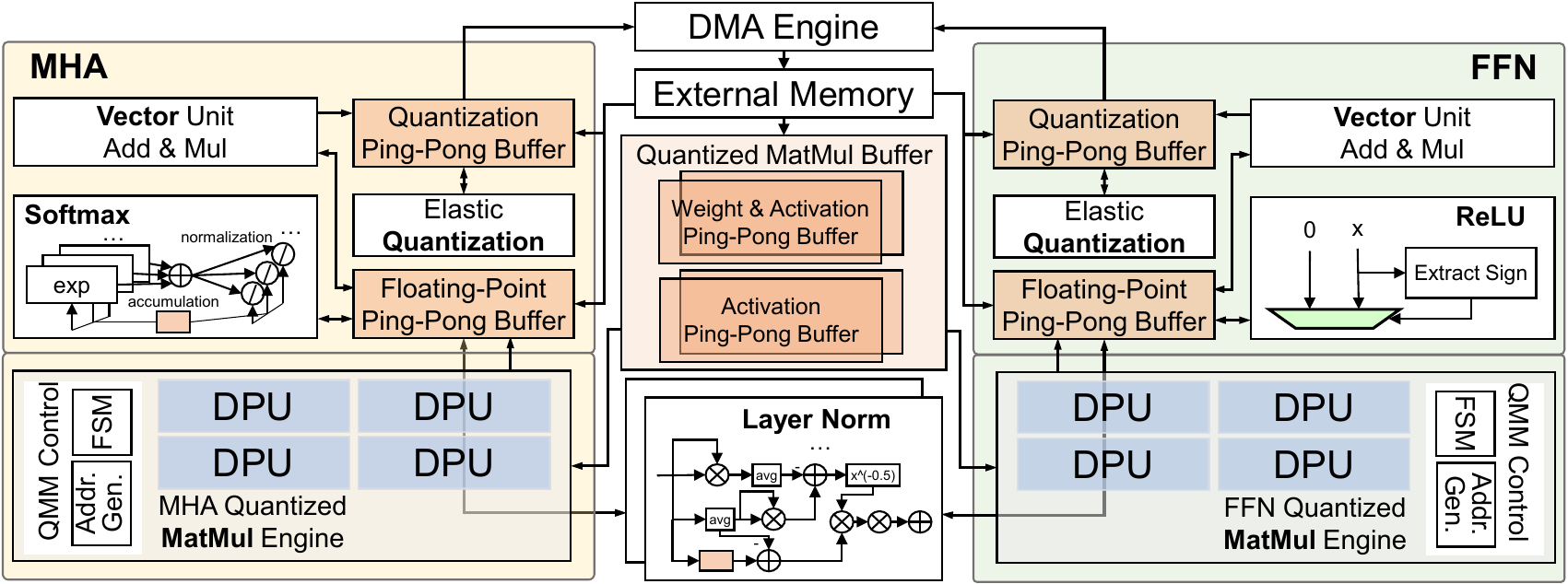}
    \caption{Hardware overview of our proposed BAT.}
    \label{fig:arch}
    \vspace{-0.4cm}
\end{figure*}

\subsection{Weighted Ternary Weight Splitting} \label{subsec:WTWS}

\fc{To compensate for potential accuracy loss from hardware-oriented design choices, BMT introduces a novel technique called weighted ternary weight splitting (WTWS). 
Unlike prior techniques like TWS in BinaryBERT~\cite{BinaryBERT}, WTWS leverages trainable coefficients during the splitting process. These coefficients capture the importance of each splitted weight, leading to higher accuracy.}


As shown in \Cref{fig:wtws}, TWS starts from training a half-sized ternary Transformer model, and then applies ternary weight splitting operator on its linear layers to obtain the full-sized latent binarized Transformer model, which quantizes weights to \{-1, 1\}. The equations are as follows:


\begin{equation}
\begin{aligned}
& W_{1, i}^b, W_{2, i}^b = \begin{cases}a \cdot W_i^t, (1-a) W_i^t & \text { if } \hat{W}_i^t \neq 0, \\
b+W_i^t, -b & \text { if } \hat{W}_i^t=0, W_i^t>0, \\
b, -b+W_i^t & \text { otherwise, }\end{cases} 
\end{aligned}
\end{equation}
where $a$ and $b$ are defined as follows:
\begin{equation}
\begin{footnotesize}
\begin{aligned}
a=\frac{\sum_{i \in \mathcal{I}}\left|W_i^t\right|+\sum_{j \in \mathcal{J}}\left|W_j^t\right|-\sum_{k \in \mathcal{K}}\left|W_k^t\right|}{2 \sum_{i \in \mathcal{I}}\left|W_i^t\right|}, b=\frac{\frac{n}{|\mathcal{I}|} \sum_{i \in \mathcal{I}}\left|W_i^t\right|-\sum_{i=1}^n\left|W_i^t\right|}{2(|\mathcal{J}|+|\mathcal{K}|)},
\end{aligned}
\end{footnotesize}
\end{equation}
where $\mathcal{I} = \{i ~|~ \hat{W}_i^t \neq 0\}$, $\mathcal{J} = \{j ~|~ \hat{W}_j^t = 0 ~ and ~ W_j^t \textgreater 0 \}$, $\mathcal{K} = \{k ~|~ \hat{W}_k^t = 0 ~ and ~ W_k^t \textless 0\}$. $\left| \cdot \right|$ denotes the cardinality of the set. The equation above ensures the result of the linear layer remains the same, i.e. $\hat{W}^t = \hat{W}_1^b + \hat{W}_2^b$. 

\fc{However, simply adding the split weights in TWS may discard valuable information.}
For example, when $\hat{W}_1^b \gg \hat{W}_2^b$, $\hat{W}^t \approx \hat{W}_1^b$ and the information of  $\hat{W}_2^b$ may be lost. 
To alleviate this problem, 
WTWS introduces trainable coefficients $\sigma_1$ and $\sigma_2$ to learn the significance of $\hat{W}_1^b$ and $\hat{W}_2^b$ respectively, reformulating the equation to $Y  = X(\hat{W}^b_1 \cdot \sigma_1 + \hat{W}^b_2 \cdot \sigma_2)$. 
As shown in \Cref{fig:wtws}, WTWS introduces an additional layer highlighted in orange to facilitate this process.
Both $\sigma_1$ and $\sigma_2$ are initialized as all-ones vectors and are column-wise trainable.

\section{Hardware Accelerator}\label{sec:accel}

\update{Considering the disadvantages of primitive streaming or processor-like accelerators, our proposed binarized Transformer accelerator, namely BAT, is characterized by a streaming-processor-mixed architecture.}
\update{On the one hand, the MHA and FFN are executed separately in a pipelined manner following a streaming-like design.
On the other hand, the architectures within MHA and FFN modules are designed as a processor-like paradigm.}

\subsection{Architecture Overview}
\shk{\Cref{fig:arch} shows the architecture of BAT, which is composed of a MHA module, a FFN module, a direct memory access (DMA) engine, an external memory, and several on-chip buffers.}
Both the MHA and FFN modules are equipped with a QMM engine, a vector unit (VU), a elastic quantization unit (QU) and a layer normalization unit. The MHA module includes a softmax unit while the FFN module features a ReLU activation unit.
\shk{The individual design of MHA and FFN modules significantly simplifies the control logic and datapath compared to a simple processor-like architecture where the MHA and FFN are processed 
\Rone{within a unified architecture \update{\cite{STA, Lu, jialin2023pptransformer-iccad}}.}
}

\Rone{In both MHA and FFN modules, the dominant operations
, QMMs, are performed by dot-product-based QMM engine which can flexibly adapt to different matrix multiplication sizes with high efficiency.}
\shk{VU handles operations with low computational density including dequantization and residual addition with vectorized inputs, and QU is designed for elastic quantization.
To ease the on-chip memory consumption and avoid the overhead associated with pattern matching between the outputs of QU and the inputs of QMM engine, these intermediate results are temporarily transferred back to the external memory.
\update{Despite increasing the bandwidth requirement, it is a strategic trade-off that conserves valuable on-chip resources.}
Besides, ping-pong buffer technique is also implemented for all the on-chip buffers to overlap the time between data transfer and computation.}

\begin{figure}[bht]
    \centering
    \includegraphics[width=0.46\textwidth]{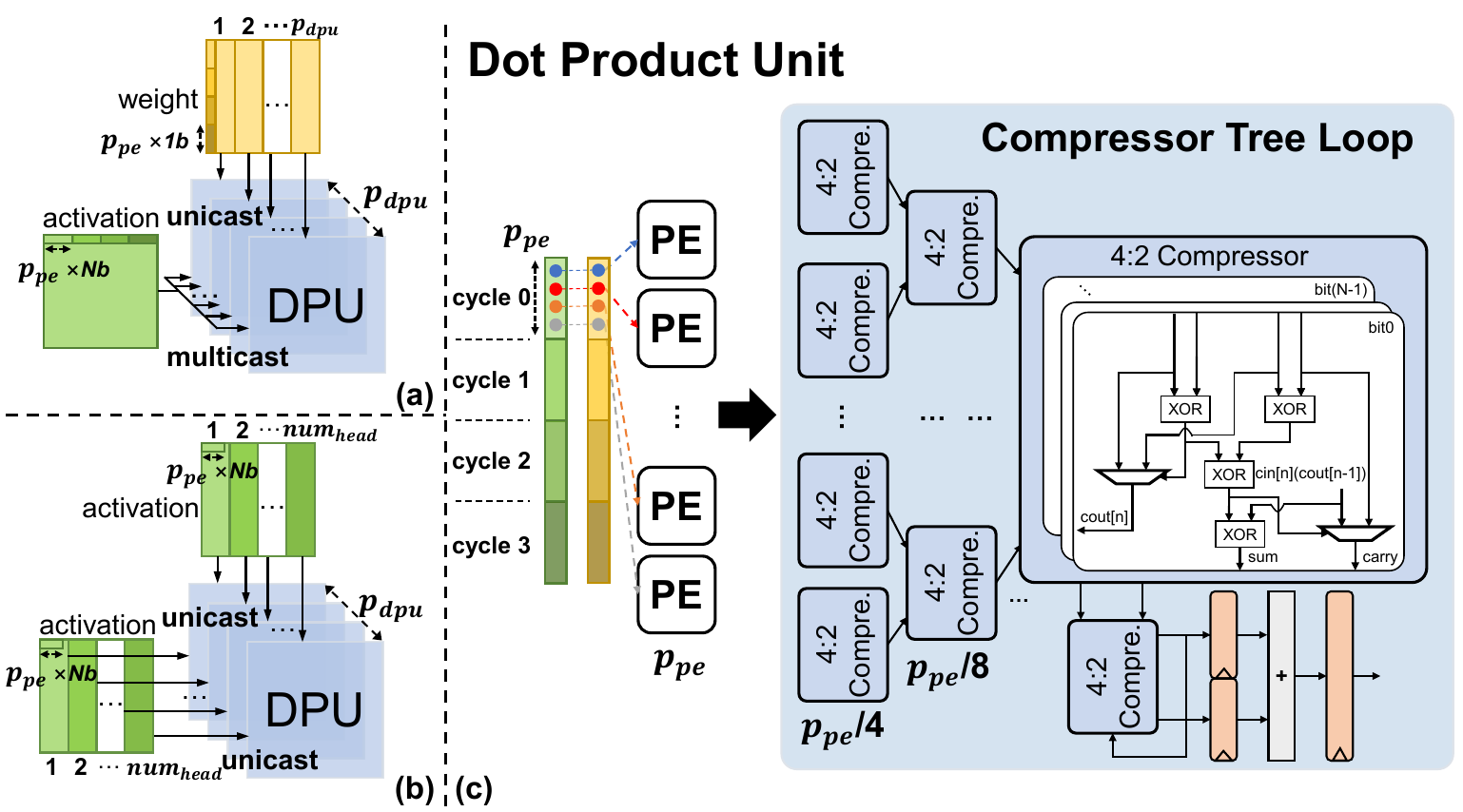}
    \caption{Data access pattern for (a) $\textit{$activation \times weight$}$ and (b) $\textit{$activation \times activation$}$ QMMs. (c) The structure of DPU.}
    \label{fig:dpu}
    \vspace{-0.4cm}
\end{figure}

\subsection{Quantized Matrix Multiplication engine}\label{subsec:QMMengine}

\fc{Processor-like architectures hinge on the core computation engine for overall performance, necessitating a QMM engine designed for high throughput and flexibility to support diverse QMMs.}
\shk{\Cref{fig:arch} shows the hardware architecture of QMM engine, which consists of $p_{dpu}$ number of dot product units (DPUs) \fc{with controllers}
including a finite state machine (FSM) and an address generator.
}

\shk{
As mentioned in \Cref{subsec:quan}, there are two types of QMMs in the quantized Transformer: (a) \textit{$activation \times weight$} and (b) \textit{$activation \times activation$}.}
\shk{\Cref{fig:dpu} (a) and (b) illustrates the unified data access pattern of QMM engine for these two QMMs:}
(a) The weight matrix are partitioned into $p_{dpu}$ tiles along the column direction, with each tile being fed into the corresponding DPU. Concurrently, the activation matrix is multicast to each DPU.
\shk{(b) QMM follows the multi-head mechanism, both activation matrices are partitioned into $num_{head}$ tiles along the column direction and are then unicast to their respective DPU.
When $num_{head}$ and $p_{dpu}$ are not exactly matched, the $num_{head}$ tiles are batched ($num_{head} > p_{dpu}$) or divided along the row dimension ($num_{head} < p_{dpu}$).}
\shk{Notably, the MHA's QMM engine is designed to incorporate both patterns while the FFN's QMM engine only needs pattern (a), as the FFN contains only the \textit{$activation \times weight$} QMMs.}


\fc{\Cref{fig:dpu} (c) elaborates on the structure of DPU.}
\shk{Each DPU is composed of $p_{pe}$\fc{-parallel} processing element units (PEs), followed by a compressor tree loop.}
\fc{For high throughput, the DPU executes vector dot products in an unfolding factor of $p_{pe}$, processing $p_{pe}$ elements per vector simultaneously. The compressor tree loop accumulates PE outputs in each iteration.}
The compressor tree is constructed based on 4:2 compressor, which mitigates the delay of carry chain propagation and thus optimizes the combinational delay.

\begin{table}[hb]
  \caption{Operands \& \fc{Operations} in Binarized Transformers\label{tab:x-y}}
  \centering
  \begin{tabular}{l|l|l}
  \hline
  $\boldsymbol{x}$ (\textit{activation}) & $\boldsymbol{y}$ (\textit{weight/activation}) &\textbf{\fc{Operations}}\\
  \hline
  $N$-bit signed &  binarized weight & \textbf{AW}, \textbf{BW}, \textbf{CW} \\
  $N$-bit signed & $N$-bit signed activation  & \textbf{Q}$\textbf{K}^T$  \\
  $N$-bit unsigned & $N$-bit signed activation  & \textbf{SV} \\
  $N$-bit unsigned &  binarized weight & \textbf{R1W} \\
  \hline
  \end{tabular}
  \vspace{-0.5cm}
\end{table}

\subsection{Processing Element Unit}
\begin{figure}[ht]
    \centering
    \includegraphics[width=0.46\textwidth]{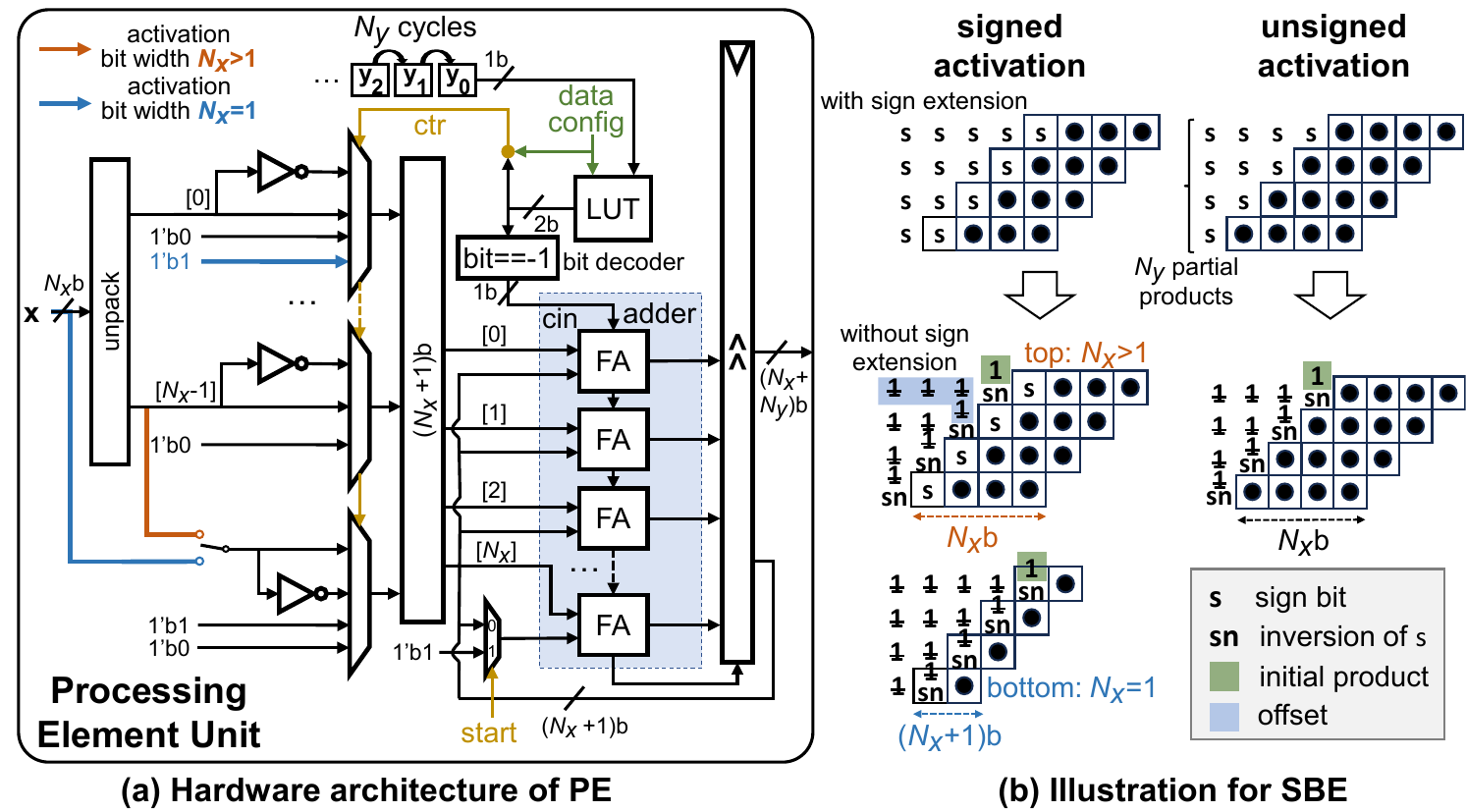}
    \caption{Detailed structure of PE and its SBE approach.}
    \label{fig:PE}
    \vspace{-0.3cm}
\end{figure}

\Cref{fig:PE} (a) depicts the detailed architecture of PE for a multiplication $\boldsymbol{x} \times \boldsymbol{y}$. 
\shk{\Cref{tab:x-y} summarizes the operands assigned to PE for different QMMs in binarized Transformer.}
To improve hardware utilization, PE is implemented with a bit-serial design. 
Each cycle, PE generates an $N$-bit $\times$ $1$-bit partial product. 
\shk{Consequently, it accomplishes $N$-bit activation $\times$ binarized weight and $N$-bit $\times$ $N$-bit activation multiplications with 1 cycle and $N$ cycles, respectively, \fc{maximizing the hardware utilization}.}

\shk{Formally, the decimal value of binary encoded $\boldsymbol{y}$ can be obtained through $\sum_{i=0}^{N_y-1}y_i2^i$, where the bit value of $y_i$ is either $1$, $0$, $-1$.}
\shk{Based on this, we develop a bit decoder to decode $y_i$ into its real bit value (1: 2'b01, 0: 2'b00, -1: 2'b11) according to its data config, which can be implemented as a 2-bit output look-up table (LUT).}

We then introduce a \shk{novel} Sign Bit Elimination (SBE) approach to preprocess the signed or unsigned \Rone{$N_x$-bit activation $\boldsymbol{x}$}, aiming to save the \shk{resource consumption} and shorten the critical path of PE. 
As illustrated in \Cref{fig:PE} (b), the multiplication unfolds into a summation of $N_y$ partial products $\boldsymbol{p}=\boldsymbol{x} \times y_i$, represented in two's complement. 
\Rone{
In the case where $\boldsymbol{x}$ is signed and $N_x \textgreater 1$, referred to as the top case, the \shk{traditional extended sign bits} of each partial product are reformulated to an addition of the inversion of sign bit ($sn$) and multiple bit values of $1$.
Here, most of the bit values of $1$ are offset, with only one left, termed initial product, and then the summation is transformed to a similar structure of unsigned integer addition without sign extension, with the bit width of each partial product increasing to $(N_x+1)$.
}
Similarly, in the case where $\boldsymbol{x}$ is unsigned, SBE initiates from the sign bit $p_{N_x}$ and the bit width also increases to $(N_x+1)$. 
\update{When $N_x=1$ and $\boldsymbol{p}$ $(-1/0/1)$ requires a binary representation of $N_x+1=2$ bits rather than $N_x$ bits in other cases, SBE initiates from $p_1$ and the bit width remains $N_x+1$ after SBE, which is consistent with other cases and shown in the signed bottom case of \cref{fig:PE} (b).}

\begin{figure}[ht]
    \centering
    \includegraphics[width=0.46\textwidth]{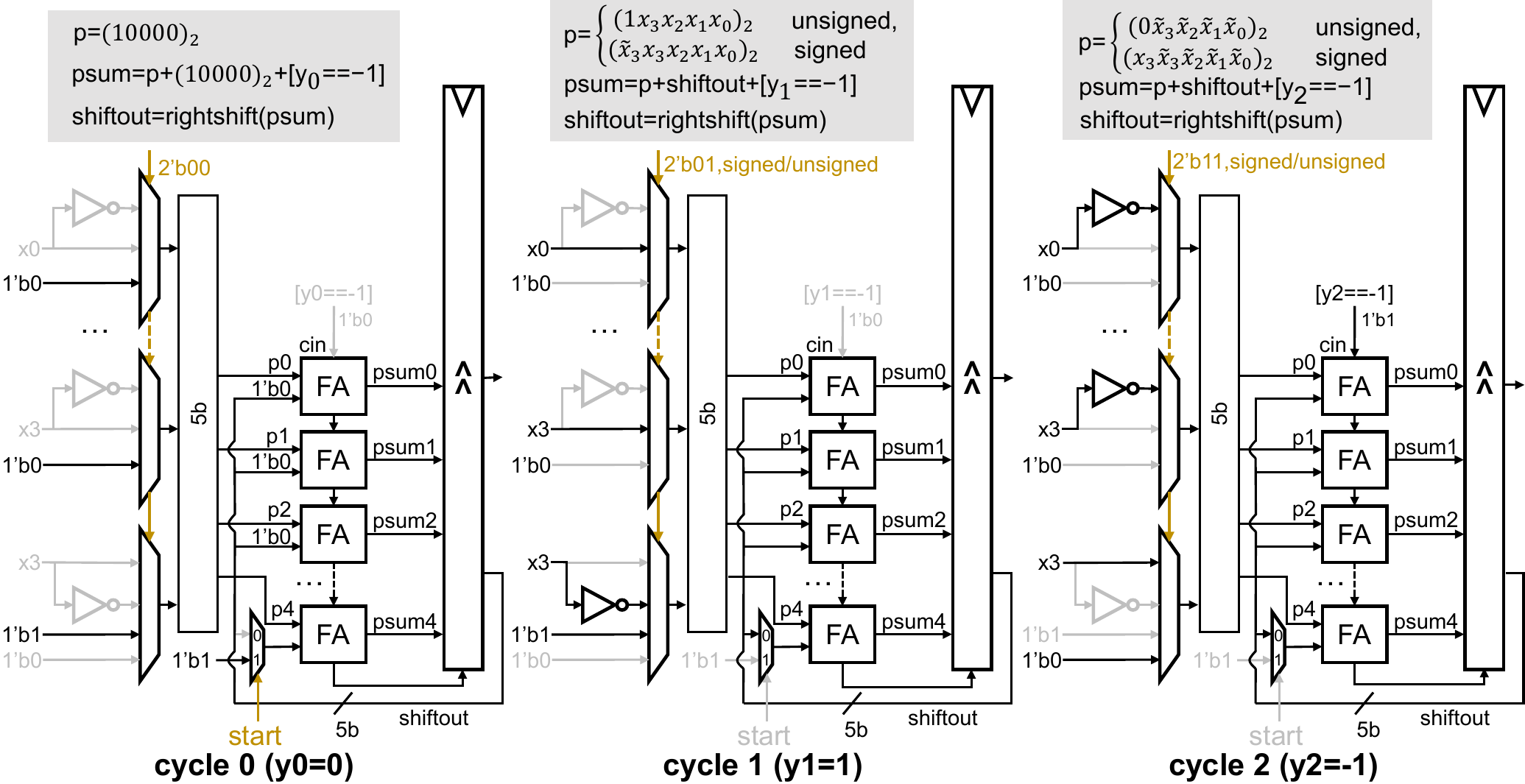}
    \caption{Example of the workflow of the proposed PE ($N_x$=4).}
    \label{fig:PEFlow}
    \vspace{-0.2cm}
\end{figure}

\shk{\Cref{fig:PEFlow} further showcases the $\boldsymbol{x} \times \{\ldots y_2y_1y_0\}$ computation process of PE with SBE when $N_x=4$.}
\textbf{First cycle}: $y_0=0$, the value of partial product $\boldsymbol{x} \times y_0$ is $(0000)_2$ and $sn$ is $1$. 
\shk{Therefore, $\boldsymbol{p}=(10000)_2$ after SBE, which is fed into the adder together with the initial product $1$. 
The partial sum ($\textit{psum}$) is then loaded into the right shift register.
\textbf{Second cycle}: $y_1=1$, the value of $\boldsymbol{x} \times y_1$ is $\boldsymbol{x}$. Following the signed top case or the unsigned case, $p_i$ is selected through the multiplexers.
\textbf{Third cycle}: $y_2=-1$, the partial product is $-\boldsymbol{x}$, which can be reformulated to $\boldsymbol{\tilde{x}}+1$. Term $\boldsymbol{\tilde{x}}$ undergos the SBE transformation and the bit value $1$ is directed to the $c_{in}$ port of the adder.}
\shk{Through SBE, PE unifies signed and unsigned QMM operations.}
\update{Also, since there is no sign extension involved, PE is implemented as a right-shifting sequential multiplier, which requires a smaller adder and prevents long carry propagation compared to a left-shifting design, thereby saving the hardware resource and reducing the path delay \cite{reviewandbenchmarking}.}

\begin{figure}[ht]
    \centering
    \includegraphics[width=0.46\textwidth]{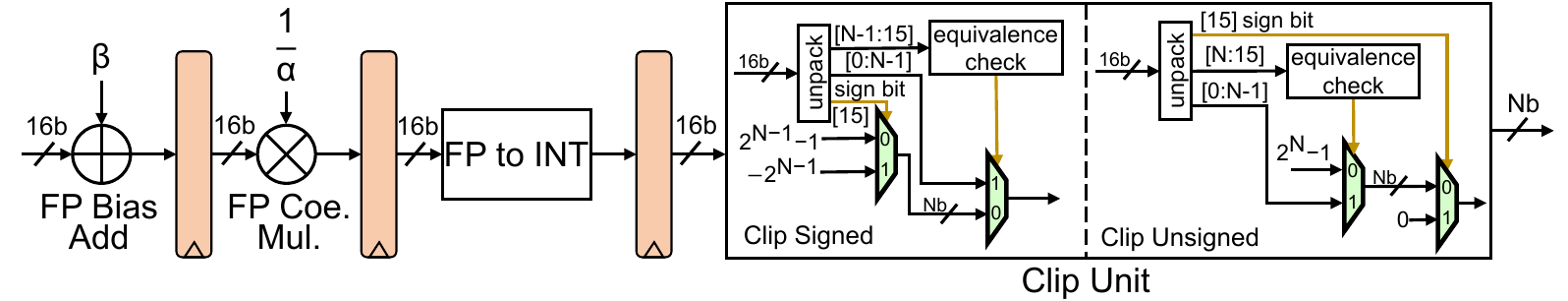}
    \caption{Elastic quantization unit.}
    \label{fig:quanunit}
    \vspace{-0.5cm}
\end{figure}

\subsection{Elastic Quantization Unit} \label{subsec:elastic}

\update{\Cref{fig:quanunit} presents our fully pipelined elastic quantization unit.}
It consists of a floating-point bias adder, a floating-point coefficient multiplier, a converter from floating-point to integer, and a clip unit. 
\fc{For specific clipping ranges like signed or unsigned numbers, the clip unit logic can be significantly simplified, as shown in \Cref{fig:quanunit}, using a combination of bit-wise operations and selection mechanisms.}
\fc{For instance, consider clipping an 16-bit signed activation to a 4-bit signed integer.}
\fc{The first step is to check if the most 13 significant bits are identical.}
\fc{Identical bits indicate that the input falls within the representable range of a 4-bit signed number and can be directly output. Conversely, non-identical high bits suggest the input is outside this range.}
In this case, the sign bit is checked and the output is set to $2^{(4-1)}-1$ if the sign bit is 0, or $-2^{(4-1)}$ if the sign bit is 1. The logic for clipping unsigned integers is similar.

\begin{figure}[htbp]
    \centering
    \includegraphics[width=0.44\textwidth]{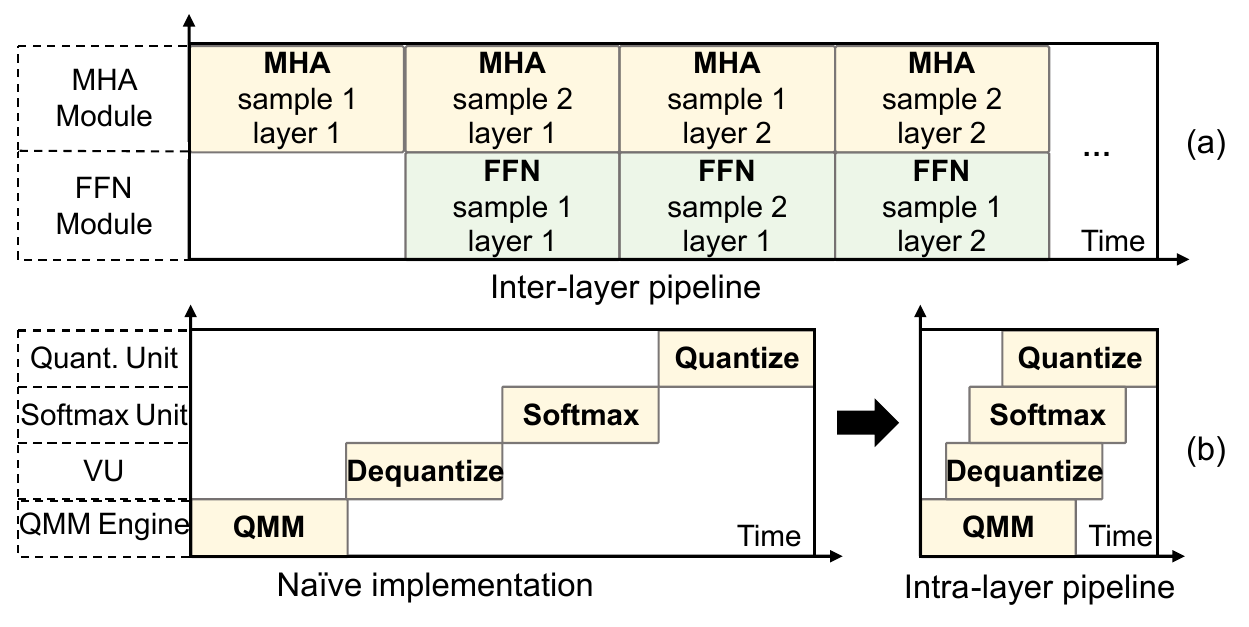}
    \caption{Dataflow optimization: (a) inter-layer pipeline and (b) intra-layer pipeline.}
    \label{fig:dataflow}
    \vspace{-0.5cm}
\end{figure}

\section{Scheduling and Co-optimization} \label{sec:opt}
 
\subsection{Dataflow Optimization} \label{subsec:dataflow}
We propose two dataflow optimization methods, inter-layer pipeline, and intra-layer pipeline, for the streaming-like and processor-like design in BAT, respectively, \fc{enhancing the processing efficiency.}

\minisection{Inter-Layer Pipeline.} 
\wx{We propose an inter-layer pipeline to fully utilize the benefits of the streaming-like design with individual computational modules of MHA and FFN.
As shown in \Cref{fig:dataflow} (a), in the first stage, the MHA module loads input data and performs computations for sample 1. 
Once this process is finished, the MHA module proceeds to compute the next sample in the second stage, while the FFN module handles the computations for the output results generated by the MHA module in the first stage. 
In the third stage, the FFN module sends the output activations to the MHA module for the computations of layer 2 in sample 1.
This alternating process enables the MHA and FFN modules to perform computations for multiple layers in both samples without encountering pipeline bubbles.
Besides, as we set the batch size to 2 in this case, it is suitable for the mini-batch processing in edge scenarios.
The overall latency can be given by:}
\begin{equation}
\begin{aligned}
    T_{total} = T_{MHA} + (L-1) \times max(T_{MHA}, T_{FFN}) + T_{FFN},
\end{aligned}
\end{equation}
where $L$ is the number of model layers.




\minisection{Intra-Layer Pipeline.} 
\wx{Given that both \update{processor-like} MHA and FFN modules operate with a quantization-dequantization scheme, we introduce an intra-layer pipeline in a row-by-row execution manner to maximize computation overlap.
For example, the operations within QMM and softmax units are conducted in integer and floating-point arithmetic, respectively.
As shown in \Cref{fig:dataflow}(b), once the softmax unit completes the computation of the first row in an input tensor, the results are immediately fed to the quantization unit.
By employing our proposed intra-layer pipeline, BAT achieves a noticeable reduction in overall execution latency compared to the non-pipelined implementation with the elimination of waiting time for continuous processing.
Additionally, this method allows for buffering data in certain rows instead of the entire intermediate data, leading to efficient utilization of storage resources.}
\subsection{Algorithm and Hardware Co-Optimization}
\wx{The overall design space of our end-to-end system is formed by the hyperparameters of binarized Transformers and BAT, consisting of algorithmic parameters and hardware parameters in MHA and FFN modules,
\Rone{as presented in \Cref{tab:dse}}.
Since the softmax operation constitutes a relatively small fraction of the total computational process, we set the parallelism of the softmax module to 4 without further exploration.}

\begin{table}[htbp]
\centering
\caption{Design Space of Co-Optimization}
\label{tab:dse}
\resizebox{0.46\textwidth}{!}{%
\begin{threeparttable}
\begin{tabular}{cccc}
\hline
\multicolumn{4}{c}{\textbf{Algorithmic Parameters}$^{*}$} \\ \hline
\multicolumn{1}{c|}{Model Type} & \multicolumn{1}{c|}{$d_{hid}$} & \multicolumn{1}{c|}{$d_{inter}$} & $b_{act}$ \\ \hline
\multicolumn{1}{c|}{BinaryBERT, BMT, BiT} & \multicolumn{1}{c|}{192, 384, 768} & \multicolumn{1}{c|}{768, 1536, 3072} & \begin{tabular}[c]{@{}c@{}}BinaryBERT: 8, 4\\ BMT: 4, 2 \, BiT: 1\end{tabular} \\ \hline 
\multicolumn{4}{c}{\textbf{Hardware Parameters}$^{**}$} \\ \hline
\multicolumn{1}{c|}{$p_{dpu}$} & \multicolumn{1}{c|}{$p_{quan}$} & \multicolumn{1}{c|}{$p_{vu}$} & $p_{ln}$ \\ \hline
\multicolumn{1}{c|}{8, 16, 32, 48, 64, 96} & \multicolumn{1}{c|}{32, 64, 128} & \multicolumn{1}{c|}{32, 64, 96, 128, 160} & 8, 16, 32, 48 \\ \hline
\end{tabular}%
\begin{tablenotes}
\scriptsize
\item $^*$ Hidden dimension size ($d_{hid}$), intermediate dimension size ($d_{inter}$), activation quantization precision ($b_{act}$).
\item $^{**}$ Parallelism of DPUs ($p_{dpu}$), quantization unit ($p_{quan}$), vector processing unit ($p_{vu}$), and LN unit ($p_{ln}$).
\end{tablenotes}
\end{threeparttable}
}
\vspace{-0.3cm}
\end{table}

\begin{table*}[t]
\centering
\caption{Results on the GLUE Development Set}
\label{tab:eval-alg}
\resizebox{0.90\textwidth}{!}{%
\begin{threeparttable}
\begin{tabular}{ccccccccccccc}
\hline
\textbf{Quant} & \textbf{\#Bits (E-W-A)$^\dagger$} & \multicolumn{1}{l}{\textbf{Size (MB)}} & \multicolumn{1}{l}{\textbf{FLOPs (G)$^\ddagger$}} & \textbf{MNLI$_{m/mm}$} & \textbf{QQP} & \textbf{QNLI} & \textbf{SST-2} & \textbf{CoLA} & \textbf{STS-B} & \textbf{MRPC} & \textbf{RTE} & \textbf{Avg.} \\ \hline
BERT~\cite{BERT} & 32-32-32 & 418 & 22.5 & 84.9/85.5 & 91.4 & 92.1 & 93.2 & 59.7 & 90.1 & 86.3 & 72.2 & 83.9 \\ \hline
Q-BERT~\cite{QBERT} & 2-8-8 & 43.0 & 6.5 & 76.6/77.0 & - & - & 84.6 & - & - & 68.3 & 52.7 & - \\
Q2BERT~\cite{zafrir2019q8bert} & 2-8-8 & 43.0 & 6.5 & 47.2/47.3 & 67.0 & 61.3 & 80.6 & 0 & 4.4 & 68.4 & 52.7 & 47.7 \\
TernaryBERT~\cite{TernaryBERT} & 2-2-8 & 28.0 & 6.4 & 83.3/83.3 & 90.1 & - & - & 50.7 & - & 87.5 & 68.2 & - \\
BinaryBERT~\cite{BinaryBERT} & 1-1-8 & 16.5 & 3.1 & 84.2/84.7 & 91.2 & 91.5 & 92.6 & 53.4 & 88.6 & 85.5 & 72.2 & 82.7 \\
BinaryBERT~\cite{BinaryBERT} & 1-1-4 & 16.5 & 1.5 & 83.9/84.2 & 91.2 & 90.9 & 92.3 & 44.4 & 87.2 & 83.3 & 65.3 & 79.9 \\
BiT~\cite{BiT} & 1-1-4 & 13.4 & 1.5 & 83.6/84.4 & 87.8 & 91.3 & 91.5 & 42.0 & 86.3 & 86.8 & 66.4 & 79.5 \\
\textbf{BMT} & \textbf{1-1-4} & \textbf{16.8} & \textbf{1.5} & \textbf{83.2/83.3} & \textbf{91.0} & \textbf{90.4} & \textbf{92.4} & \textbf{49.7} & \textbf{83.5} & \textbf{87.5} & \textbf{67.1} & \textbf{80.9} \\
BinaryBERT~\cite{BinaryBERT} & 1-1-2 & 16.5 & 0.8 & 62.7/63.9 & 79.9 & 52.6 & 82.5 & 14.6 & 6.5 & 78.3 & 52.7 & 41.0 \\
BiT~\cite{BiT} & 1-1-2 & 13.4 & 0.8 & 82.1/82.5 & 87.1 & 89.3 & 90.8 & 32.1 & 82.2 & 78.4 & 58.1 & 75.0 \\
\textbf{BMT} & \textbf{1-1-2} & \textbf{16.8} & \textbf{0.8} & \textbf{81.2/81.5} & \textbf{90.0} & \textbf{88.3} & \textbf{91.5} & \textbf{37.4} & \textbf{71.4} & \textbf{82.1} & \textbf{61.7} & \textbf{76.1} \\
BiT~\cite{BiT} & 1-1-1 & 13.4 & 0.4 & 79.5/79.4 & 85.4 & 86.4 & 89.9 & 32.9 & 72.0 & 79.9 & 62.1 & 73.5 \\ \hline
\end{tabular}%
\begin{tablenotes}
    \item[$\dagger$] The E-W-A notation refers to the quantization bit width of embeddings, weights, and activations.
    \item[$\ddagger$] The FLOPs is calculated followed by the same approach as BinaryBERT~\cite{BinaryBERT} and the input sequence length is set to 128.
\end{tablenotes}
\end{threeparttable}
}
\vspace{-0.4cm}
\end{table*}


\wx{To achieve a globally optimized design, we assess the network deployment performance for each design variable, including accuracy, robustness, latency, and resource consumption.
Specifically, the accuracy is determined by training and evaluating the binarized Transformer. 
In terms of robustness, we introduce perturbation $\Delta x$ in the embedding layer using Gaussian noise, with a variance set to 0.01 and magnitude set to 10\% of the magnitude of its original output, similar to steps in \cite{BEBERT} and \cite{robustness}.
After repeating the inference 20 times, we obtain an accuracy array $A$. 
\update{We define the robustness as ${A_0}/\sqrt{\frac{1}{20}\sum_{i=1}^{20} (A_i-A_0)^2}$, where $A_0$ represents the reported best accuracy without perturbation.}
The end-to-end latency is derived from a performance model developed for our BAT, which is cross-validated through RTL simulation results, taking into account memory accesses to external memory.
Hardware resource consumption is obtained after synthesis to ensure the accelerator can fit within the target FPGA device. 
Finally, we employ an exhaustive grid search to explore all the design points under specific constraints, such as the minimal accuracy demand, and identify the Pareto-optimal set.}

\section{Evaluation}\label{sec:eval}

\subsection{Experimental Setup}

\minisection{Benchmarks.} We follow the experimental setting of BinaryBERT \cite{BinaryBERT} and BiT~\cite{BiT} and use the pre-trained BERT-base as our full precision baseline. 
The algorithmic performance is evaluated on the development set of GLUE \cite{GLUE}, a widely adopted benchmark across a diverse set of language understanding tasks.

\minisection{Software Implementation.} We train and test our BMT using PyTorch v1.10.1. Following BinaryBERT \cite{BinaryBERT}, we obtain the half-sized full precision models from DynaBERT~\cite{hou2020dynabert}. We adopt LSQ quantization \cite{lsq} for the half-sized ternary model and elastic quantization for the full-sized binarized model. 
Nvidia RTX 3090 GPU is used for training. 
Following BinaryBERT, the hidden dimension size and intermediate dimension size are set to 384 and 1536, respectively for our BMT, \Rone{termed full-size.}

\minisection{Hardware Implementation.} 
We implement our BAT hardware accelerator using SystemVerilog on Xilinx ZCU102 FPGA board and evaluate it for BinaryBERT \cite{BinaryBERT}, BiT \cite{BiT} and \fc{our} BMT on the MRPC task in the GLUE benchmark to obtain the end-to-end latency.
\fc{The running frequency is set to 200 MHz}, and Xilinx Vivado 2022.2 is used for synthesis and implementation. 
\fc{Power consumption values are obtained using the Xilinx Power Estimator (XPE) tool.}
\Rone{We use 16-bit half-precision floating-point (FP16) numbers to process the full precision operations in the quantization-dequantization computational flow of BMT \fc{to ensure model accuracy.}}

\subsection{Algorithmic Performance}

\fc{We compare our BMT with five existing quantized Transformer models: Q-BERT~\cite{QBERT}, Q2BERT~\cite{zafrir2019q8bert}, TernaryBERT~\cite{TernaryBERT}, BinaryBERT \cite{BinaryBERT}, and BiT \cite{BiT}, as well as the full precision pre-trained BERT as the baseline.}
\fc{\Cref{tab:eval-alg} presents the main results on the GLUE benchmark.}
Compared with SOTA binarized Transformers, our proposed BMT improves the average performance by 1.0\% and 1.1\% in $W1A4$ and $W1A2$ quantization configuration, respectively. 
\fc{Notably, compared to the full-precision baseline, our $W1A4$ BMT retains competitive performance with a significant reduction of model size and computation up to 25$\times$ and 15$\times$, respectively}, demonstrating significant acceleration potential for edge deployment.

\begin{figure}[ht]
    \centering
    \includegraphics[width=0.48\textwidth]{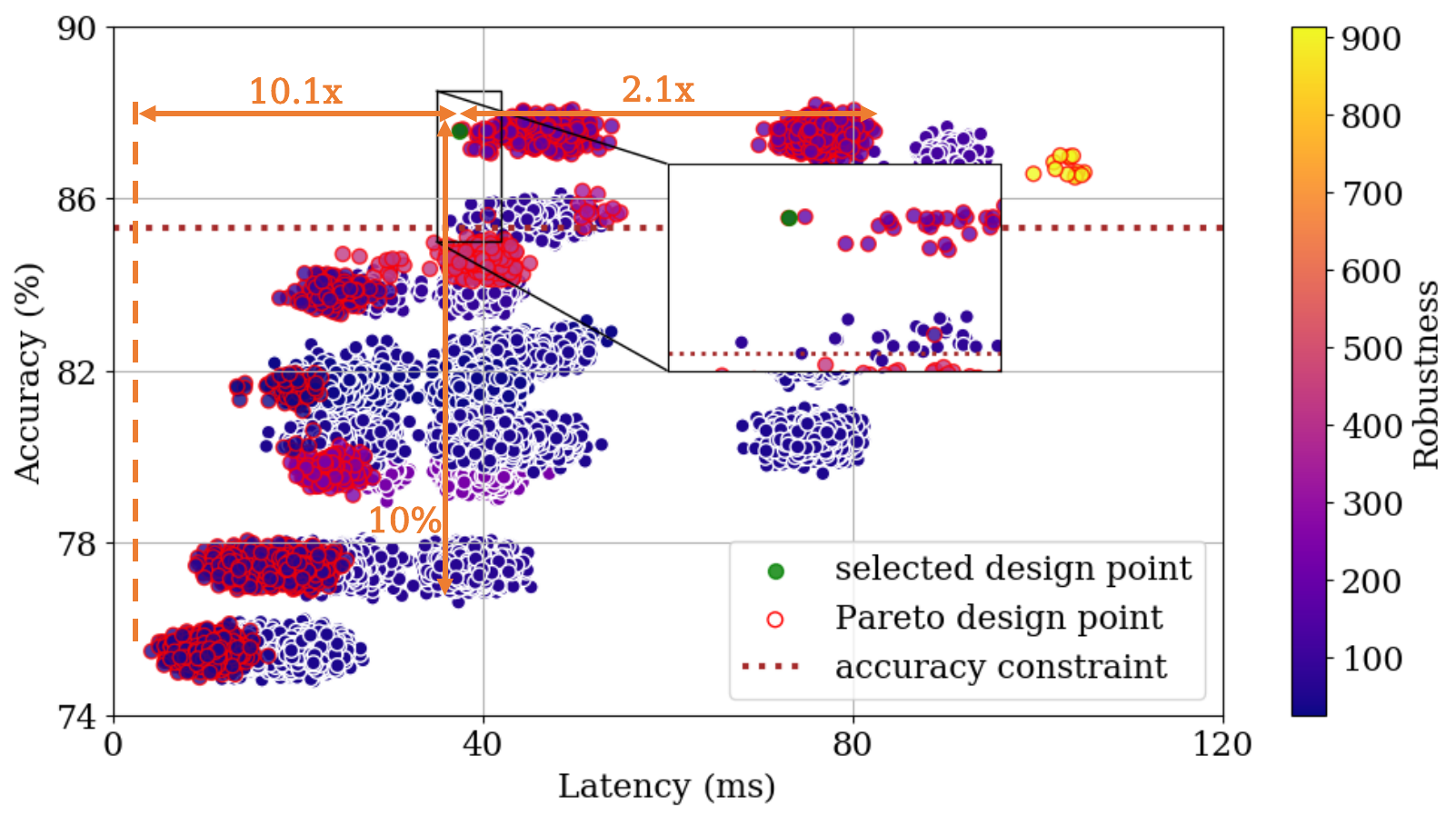}
    \caption{Design points of co-optimization on MRPC dataset.}
    \label{fig:dse}
    \vspace{-0.4cm}
\end{figure}


\begin{table*}[htbp]
\centering
\caption{Comparison of BAT with Prior Arts and Commercial Products}
\label{tab:eval_hw}
\resizebox{0.92\textwidth}{!}{%
\begin{tabular}{@{}c|cc|c|ccccc@{}}
\toprule
 & \multicolumn{2}{c|}{CPU} & GPU & \multicolumn{5}{c}{FPGA} \\ \cmidrule(l){2-9} 
 &  & {\color[HTML]{060607} } &  & FTRANS \cite{FTRANS} & STA \cite{STA} & ViA \cite{ViA} & FQ-BERT \cite{FQ-BERT} & \textbf{BAT} \\ \cmidrule(l){5-9} 
\multirow{-3}{*}{Platform} & \multirow{-2}{*}{\begin{tabular}[c]{@{}c@{}}Intel\\ i9-9900X\end{tabular}} & \multirow{-2}{*}{{\color[HTML]{060607} \begin{tabular}[c]{@{}c@{}}AMD \\ EPYC 7302\end{tabular}}} & \multirow{-2}{*}{\begin{tabular}[c]{@{}c@{}}NVIDIA \\ RTX 3090\end{tabular}} & VCU118 & XC7Z020 & Alveo U50 & ZCU102 & \textbf{ZCU102} \\ \midrule
Technology & 14nm & 7nm & 8nm & 16nm & 28nm & 16nm & 16nm & \textbf{16nm} \\ \midrule
Frequency (Hz) & 3.5G & 3.0G & 1.7G & 170M & 200M & 300M & 214M & \textbf{200M} \\ \midrule
Test Network & BMT & BMT & BMT & \begin{tabular}[c]{@{}c@{}}BCM-based \\ Compressed RoBERTa\end{tabular} & \begin{tabular}[c]{@{}c@{}}N:M Sparse\\ Shallow Transformer\end{tabular} & Swin Transformer & FQBERT & \textbf{BMT} \\ \midrule
Throughput (GOPS) & 196.73 & 148.88 & 413.10 & 101.80 & 523.81 & 309.60 & 22.74 & \textbf{1122.40} \\ \midrule
Power (W) & 165.00 & 155.00 & 350.00 & 25.06 & 9.87 & 39.00 & 9.80 & \textbf{5.47} \\ \midrule
Energy Eff. (GOPS/W) & 1.19 & 0.96 & 1.18 & 4.06 & 55.16 & 7.94 & 2.32 & \textbf{207.70} \\ \bottomrule
\end{tabular}%
}
\vspace{-0.1cm}
\end{table*}

\subsection{Effectiveness of Co-optimization}

\fc{We then evaluate the effectiveness of our co-design method in finding the Pareto-optimal configurations in both algorithm and hardware aspects.}
\fc{The MRPC dataset is used for demonstration purposes.}
\fc{The design space is presented in Table~\ref{tab:dse}, and we constrain the total hardware resource consumption to be less than 80\% of the available FPGA resources to facilitate implementation.}

\fc{\Cref{fig:dse} illustrates the trade-off between accuracy, robustness, and latency.
\Rone{The red circles indicate the Pareto optimal points.}
The brown dashed line represents the accuracy constraint, set at a maximum loss of 1\% compared to the full-precision BERT-base model. 
\Rone{We also set the robustness constraint of exceeding the upper quartile.}
\Rone{Among Pareto optimal points satisfying the constraints,}
we select the one with the lowest latency. 
The results demonstrate that our co-designed solution achieves up to 10\% higher accuracy than other design points within the same latency range and is 2.1$\times$ faster than those with similar accuracy, all while maintaining comparable robustness.}
\fc{The resulting optimal configurations for the MRPC dataset are:
\begin{itemize}
    \item <model type, $d_{hid}$, $d_{inter}$, $b_{act}$> = <BMT, 384, 1536, 4>;
    \item <$p_{dpu}$, $p_{quan}$, $p_{vu}$, $p_{ln}$> = <16, 128, 32, 8> for MHA module;
    \item <$p_{dpu}$, $p_{quan}$, $p_{vu}$, $p_{ln}$> = <16, 128, 96, 8> for FFN module.
\end{itemize}}
\fc{These configurations will be used for the evaluation in the remaining sections unless otherwise specified.}

\fc{While the selected design point is approximately 10.1$\times$ slower than the one with the absolute lowest latency, it adheres to the crucial constraints of accuracy and robustness, which are essential for real-world deployments.}
This global balance ensures that our system not only performs efficiently but also maintains high standards of reliability and \fc{accuracy}.

\subsection{Hardware Evaluation}

\subsubsection{Energy Efficiency}

\begin{figure}[ht]
    \vspace{-0.3cm}
    \centering
    \includegraphics[width=0.46\textwidth]{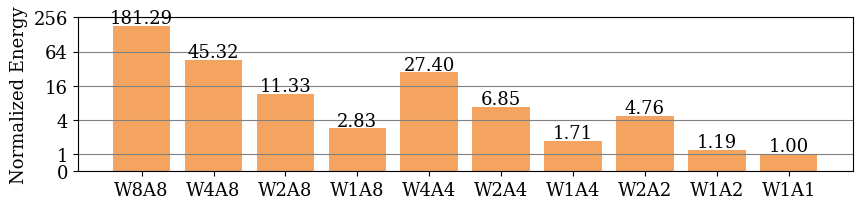}
    \caption{Normalized energy consumption for QMM under different quantization precisions.}
    \label{fig:energy}
    \vspace{-0.4cm}
\end{figure}

\fc{To explore the energy efficiency benefits of binarized Transformers}, we choose 10 commonly quantized network configurations and deploy their corresponding $activation \times weight$ QMM on our QMM Engine. 
\update{The size of QMM is set to $(128, 768) \times (768, 768)$.}
\fc{As shown in \Cref{fig:energy}, the binarized Transformer configuration, where weights are quantized to 1-bit, achieves significantly lower energy consumption compared to other configurations by 2$\sim$8$\times$ under the same activation precision.}

\subsubsection{Hardware Consumption}

\Cref{tab:breakdown} presents the FPGA resource consumption and power breakdown of BAT. 
\fc{Due to its bit-wise operations, the QMM Engine is not efficiently mapped to DSP units and primarily relies on LUT resources. Notably, the QMM Engine consumes a modest 10.90\% of the total power consumption, highlighting the effectiveness of our optimization for matrix multiplication.}
\fc{The DSP usage within the Quantization Unit stems from the floating-point bias adders and coefficient multipliers. 
Other components, mainly consisting of floating-point operations and on-chip buffer, consume 69.19\% of the total power consumption, indicating a potential bottleneck for further energy optimization.}

\begin{table}[htb]
\centering
\caption{FPGA Resource Consumption \& Power Breakdown}
\label{tab:breakdown}
\resizebox{0.48\textwidth}{!}{%
\begin{tabular}{@{}c|ccccc@{}}
\toprule
\multicolumn{1}{l|}{} & \textbf{LUT} & \textbf{FF} & \textbf{BRAM} & \textbf{DSP} & \textbf{Power (W)} \\ \midrule
\textbf{MHA QMM Engine} & \begin{tabular}[c]{@{}c@{}}37338\\ (25.49\%)\end{tabular} & \begin{tabular}[c]{@{}c@{}}13725\\ (10.02\%)\end{tabular} & - & - & \begin{tabular}[c]{@{}c@{}}0.298\\ (5.45\%)\end{tabular} \\ \midrule
\textbf{FFN QMM Engine} & \begin{tabular}[c]{@{}c@{}}37330\\ (25.49\%)\end{tabular} & \begin{tabular}[c]{@{}c@{}}13702\\ (10.00\%)\end{tabular} & - & - & \begin{tabular}[c]{@{}c@{}}0.298\\ (5.45\%)\end{tabular} \\ \midrule
\textbf{Quantization Unit} & \begin{tabular}[c]{@{}c@{}}13952\\ (9.53\%)\end{tabular} & \begin{tabular}[c]{@{}c@{}}29952\\ (21.87\%)\end{tabular} & - & \begin{tabular}[c]{@{}c@{}}256\\ (25\%)\end{tabular} & \begin{tabular}[c]{@{}c@{}}1.088\\ (19.01\%)\end{tabular} \\ \midrule
\textbf{Others} & \begin{tabular}[c]{@{}c@{}}57845\\ (39.49\%)\end{tabular} & \begin{tabular}[c]{@{}c@{}}79594\\ (58.11\%)\end{tabular} & \begin{tabular}[c]{@{}c@{}}114\\ (100.00\%)\end{tabular} & \begin{tabular}[c]{@{}c@{}}768\\ (75\%)\end{tabular} & \begin{tabular}[c]{@{}c@{}}3.781\\ (69.19\%)\end{tabular} \\ \midrule
\textbf{Total} & \begin{tabular}[c]{@{}c@{}}146465\\ (100.00\%)\end{tabular} & \begin{tabular}[c]{@{}c@{}}136973\\ (100.00\%)\end{tabular} & \begin{tabular}[c]{@{}c@{}}114\\ (100.00\%)\end{tabular} & \begin{tabular}[c]{@{}c@{}}1024\\ (100.00\%)\end{tabular} & \begin{tabular}[c]{@{}c@{}}5.465\\ (100.00\%)\end{tabular} \\ \midrule
\textbf{Available} & 274080 & 548160 & 912 & 2520 & - \\ \bottomrule
\end{tabular}%
}
\vspace{-0.3cm}
\end{table}

\subsubsection{Overall System Evaluation}

\Cref{tab:eval_hw} compares our BAT with SOTA FPGA-based \Rone{end-to-end} Transformer accelerators, as well as commercial CPU and GPU products, in terms of throughput and energy consumption. 
\Rone{To adapt to the mini-batch feature at the edge, the batch size on CPU and GPU is set to 2. 
The quantization-dequantization scheme is adopted on GPU, where the QMM is implemented in high-performance CUDA kernel.}
\fc{Our proposed BAT significantly outperforms other FPGA-based accelerators, achieving 2.14$\sim$49.37$\times$ improvement in throughput and 3.72$\sim$88.53$\times$ improvement in energy efficiency. 
Compared to CPU and GPU implementations, BAT demonstrates up to 7.54$\times$ and 2.72$\times$ speedup, and 213.82$\times$ and 174.01$\times$ improvement in energy efficiency, respectively.}

\begin{figure}[ht]
    \centering
    \vspace{-0.3cm}
    \includegraphics[width=0.46\textwidth]{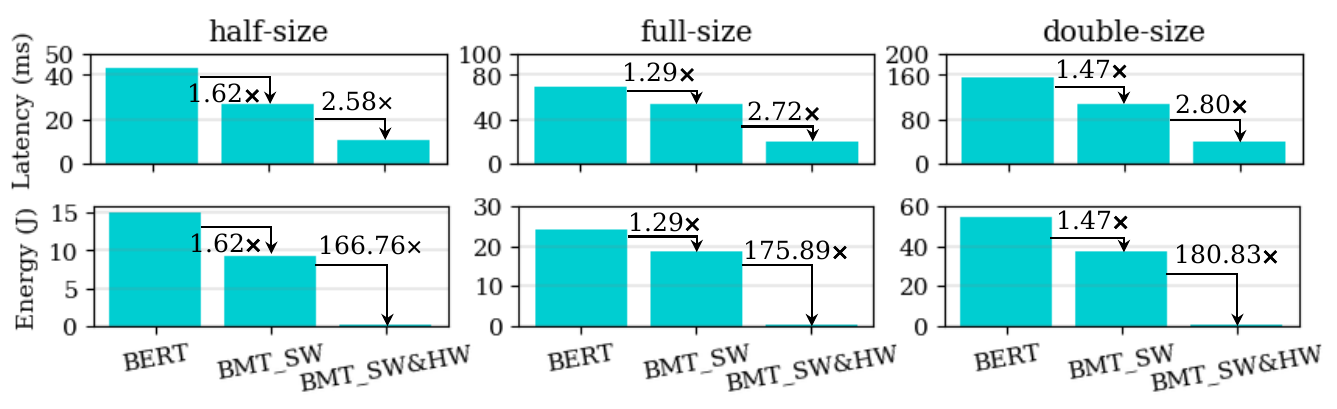}
    \caption{Speedup and energy reduction of algorithm and hardware optimizations.}
    \label{fig:ablation}
    \vspace{-0.6cm}
\end{figure}

\subsection{Ablation Study}

\fc{\Cref{fig:ablation} evaluates the contributions of our algorithm and hardware design via an ablation study.}
\Rone{We first evaluate both BERT-base and BMT under three configurations, i.e. half-size, full-size, and double-size on the NVIDIA RTX 3090 GPU to validate the performance improvement brought by our algorithm.}
\Rone{The results demonstrate a 1.29$\sim$1.62$\times$ speedup and energy savings.}
\fc{We further evaluate the BMT model on our BAT accelerator under the same configuration. 
Compared to the optimized GPU implementation, BAT achieves a 2.58$\sim$2.80$\times$ speedup and a 166.76$\sim$180.83$\times$ energy reduction.
}
\Rone{These results demonstrate the effectiveness of our co-design approach.}
\section{Conclusion} \label{sec:conclu}

The paper proposes the end-to-end acceleration of binarized Transformer via algorithm-hardware co-design \fc{enabling efficient edge deployment.}
From algorithm aspects, we propose BMT, a hardware-friendly binarized Transformer model \fc{that achieves} a substantial compression ratio while maintaining high accuracy compared to the full precision baseline.
From hardware aspects, we propose BAT, a streaming-processor-mixed binarized Transformer accelerator equipped with highly optimized computational units and dataflows, which enables efficient end-to-end inference for binarized Transformers. 
Moreover, algorithm and hardware design parameters are jointly optimized to push the performance boundaries \fc{under real-world constraints from accuracy, latency, and robustness}.
\fc{Experimental results show our co-design yields up to 2.14$\sim$49.37$\times$ 3.72$\sim$88.53$\times$ improvement on throughput and energy efficiency, respectively, over the state-of-the-art Transformer accelerators.}

\begin{acks}
This work was supported in part by the National Key R\&D Program of China under Grant 2022YFB4400600 and in part by the National Natural Science Foundation of China under Grant 62174084. The authors would like to thank Dr. Xiao Wu and Wendong Xu for their support and help.
\end{acks}

\bibliographystyle{ACM-Reference-Format-num}
\bibliography{sample}


\begin{thebibliography}{43}


\ifx \showCODEN    \undefined \def \showCODEN     #1{\unskip}     \fi
\ifx \showDOI      \undefined \def \showDOI       #1{#1}\fi
\ifx \showISBNx    \undefined \def \showISBNx     #1{\unskip}     \fi
\ifx \showISBNxiii \undefined \def \showISBNxiii  #1{\unskip}     \fi
\ifx \showISSN     \undefined \def \showISSN      #1{\unskip}     \fi
\ifx \showLCCN     \undefined \def \showLCCN      #1{\unskip}     \fi
\ifx \shownote     \undefined \def \shownote      #1{#1}          \fi
\ifx \showarticletitle \undefined \def \showarticletitle #1{#1}   \fi
\ifx \showURL      \undefined \def \showURL       {\relax}        \fi
\providecommand\bibfield[2]{#2}
\providecommand\bibinfo[2]{#2}
\providecommand\natexlab[1]{#1}
\providecommand\showeprint[2][]{arXiv:#2}

\bibitem[\protect\citeauthoryear{Bai, Zhang, et~al\mbox{.}}{Bai et~al\mbox{.}}{2021}]%
        {BinaryBERT}
\bibfield{author}{\bibinfo{person}{Haoli Bai}, \bibinfo{person}{Wei Zhang}, {et~al\mbox{.}}} \bibinfo{year}{2021}\natexlab{}.
\newblock \showarticletitle{{BinaryBERT:} Pushing the Limit of {BERT} Quantization}. In \bibinfo{booktitle}{\emph{Proceedings of the 59th Annual Meeting of the Association for Computational Linguistics and the 11th International Joint Conference on Natural Language Processing (ACL/IJCNLP)}}. \bibinfo{publisher}{ACL}, \bibinfo{pages}{4334--4348}.
\newblock


\bibitem[\protect\citeauthoryear{Brown, Mann, et~al\mbox{.}}{Brown et~al\mbox{.}}{2020}]%
        {GPT3}
\bibfield{author}{\bibinfo{person}{Tom~B. Brown}, \bibinfo{person}{Benjamin Mann}, {et~al\mbox{.}}} \bibinfo{year}{2020}\natexlab{}.
\newblock \showarticletitle{Language Models are Few-Shot Learners}. In \bibinfo{booktitle}{\emph{Advances in neural information processing systems (NeurIPS)}}.
\newblock


\bibitem[\protect\citeauthoryear{Camus, Mei, et~al\mbox{.}}{Camus et~al\mbox{.}}{2019}]%
        {reviewandbenchmarking}
\bibfield{author}{\bibinfo{person}{Vincent Camus}, \bibinfo{person}{Linyan Mei}, {et~al\mbox{.}}} \bibinfo{year}{2019}\natexlab{}.
\newblock \showarticletitle{Review and Benchmarking of Precision-Scalable Multiply-Accumulate Unit Architectures for Embedded Neural-Network Processing}.
\newblock \bibinfo{journal}{\emph{{IEEE} J. Emerg. Sel. Topics Circuits Syst.}} \bibinfo{volume}{9}, \bibinfo{number}{4} (\bibinfo{year}{2019}), \bibinfo{pages}{697--711}.
\newblock


\bibitem[\protect\citeauthoryear{Cao, Lin, et~al\mbox{.}}{Cao et~al\mbox{.}}{2023}]%
        {jialin2023pptransformer-iccad}
\bibfield{author}{\bibinfo{person}{Jialin Cao}, \bibinfo{person}{Xuanda Lin}, {et~al\mbox{.}}} \bibinfo{year}{2023}\natexlab{}.
\newblock \showarticletitle{PP-Transformer: Enable Efficient Deployment of Transformers Through Pattern Pruning}. In \bibinfo{booktitle}{\emph{{IEEE/ACM} International Conference on Computer Aided Design (ICCAD)}}. \bibinfo{publisher}{{IEEE}}, \bibinfo{pages}{1--9}.
\newblock


\bibitem[\protect\citeauthoryear{Dettmers, Lewis, et~al\mbox{.}}{Dettmers et~al\mbox{.}}{2022}]%
        {tim2022llmint8}
\bibfield{author}{\bibinfo{person}{Tim Dettmers}, \bibinfo{person}{Mike Lewis}, {et~al\mbox{.}}} \bibinfo{year}{2022}\natexlab{}.
\newblock \showarticletitle{LLM.int8(): 8-bit Matrix Multiplication for Transformers at Scale}.
\newblock \bibinfo{journal}{\emph{CoRR}}  \bibinfo{volume}{abs/2208.07339} (\bibinfo{year}{2022}).
\newblock


\bibitem[\protect\citeauthoryear{Devlin, Chang, et~al\mbox{.}}{Devlin et~al\mbox{.}}{2019}]%
        {BERT}
\bibfield{author}{\bibinfo{person}{Jacob Devlin}, \bibinfo{person}{Ming{-}Wei Chang}, {et~al\mbox{.}}} \bibinfo{year}{2019}\natexlab{}.
\newblock \showarticletitle{{BERT:} Pre-training of Deep Bidirectional Transformers for Language Understanding}. In \bibinfo{booktitle}{\emph{{Proceedings of the North American Chapter of the Association for Computational Linguistics: Human Language Technologies (NAACL-HLT)} {(1)}}}. \bibinfo{publisher}{Association for Computational Linguistics}, \bibinfo{pages}{4171--4186}.
\newblock


\bibitem[\protect\citeauthoryear{Dosovitskiy, Beyer, et~al\mbox{.}}{Dosovitskiy et~al\mbox{.}}{2021}]%
        {ViT}
\bibfield{author}{\bibinfo{person}{Alexey Dosovitskiy}, \bibinfo{person}{Lucas Beyer}, {et~al\mbox{.}}} \bibinfo{year}{2021}\natexlab{}.
\newblock \showarticletitle{An Image is Worth 16x16 Words: Transformers for Image Recognition at Scale}. In \bibinfo{booktitle}{\emph{International Conference on Learning Representations (ICLR)}}.
\newblock


\bibitem[\protect\citeauthoryear{Esser, McKinstry, et~al\mbox{.}}{Esser et~al\mbox{.}}{2020}]%
        {lsq}
\bibfield{author}{\bibinfo{person}{Steven~K. Esser}, \bibinfo{person}{Jeffrey~L. McKinstry}, {et~al\mbox{.}}} \bibinfo{year}{2020}\natexlab{}.
\newblock \showarticletitle{Learned Step Size quantization}. In \bibinfo{booktitle}{\emph{International Conference on Learning Representations (ICLR)}}. \bibinfo{publisher}{OpenReview.net}.
\newblock


\bibitem[\protect\citeauthoryear{Fan, Chau, et~al\mbox{.}}{Fan et~al\mbox{.}}{2022}]%
        {butterfly}
\bibfield{author}{\bibinfo{person}{Hongxiang Fan}, \bibinfo{person}{Thomas Chau}, {et~al\mbox{.}}} \bibinfo{year}{2022}\natexlab{}.
\newblock \showarticletitle{Adaptable Butterfly Accelerator for Attention-based NNs via Hardware and Algorithm Co-design}. In \bibinfo{booktitle}{\emph{{IEEE/ACM} International Symposium on Microarchitecture (MICRO)}}. \bibinfo{publisher}{{IEEE}}, \bibinfo{pages}{599--615}.
\newblock


\bibitem[\protect\citeauthoryear{Fang, Zhou, et~al\mbox{.}}{Fang et~al\mbox{.}}{2022}]%
        {STA}
\bibfield{author}{\bibinfo{person}{Chao Fang}, \bibinfo{person}{Aojun Zhou}, {et~al\mbox{.}}} \bibinfo{year}{2022}\natexlab{}.
\newblock \showarticletitle{An Algorithm-Hardware Co-Optimized Framework for Accelerating {N:M} Sparse {Transformers}}.
\newblock \bibinfo{journal}{\emph{IEEE Transactions on Very Large Scale Integration (VLSI) Systems (TVLSI)}} \bibinfo{volume}{30}, \bibinfo{number}{11} (\bibinfo{year}{2022}), \bibinfo{pages}{1573--1586}.
\newblock


\bibitem[\protect\citeauthoryear{Ham, Jung, et~al\mbox{.}}{Ham et~al\mbox{.}}{2020}]%
        {A3}
\bibfield{author}{\bibinfo{person}{Tae~Jun Ham}, \bibinfo{person}{Sungjun Jung}, {et~al\mbox{.}}} \bibinfo{year}{2020}\natexlab{}.
\newblock \showarticletitle{A\({}^{\mbox{3}}\): Accelerating Attention Mechanisms in Neural Networks with Approximation}. In \bibinfo{booktitle}{\emph{IEEE International Symposium on High Performance Computer Architecture (HPCA)}}. \bibinfo{publisher}{{IEEE}}, \bibinfo{pages}{328--341}.
\newblock


\bibitem[\protect\citeauthoryear{Hou, Huang, et~al\mbox{.}}{Hou et~al\mbox{.}}{2020}]%
        {hou2020dynabert}
\bibfield{author}{\bibinfo{person}{Lu Hou}, \bibinfo{person}{Zhiqi Huang}, {et~al\mbox{.}}} \bibinfo{year}{2020}\natexlab{}.
\newblock \showarticletitle{DynaBERT: Dynamic BERT with Adaptive Width and Depth}. In \bibinfo{booktitle}{\emph{Advances in Neural Information Processing Systems (NeurIPS)}}.
\newblock


\bibitem[\protect\citeauthoryear{Hua, Li, et~al\mbox{.}}{Hua et~al\mbox{.}}{2021}]%
        {robustness}
\bibfield{author}{\bibinfo{person}{Hang Hua}, \bibinfo{person}{Xingjian Li}, {et~al\mbox{.}}} \bibinfo{year}{2021}\natexlab{}.
\newblock \showarticletitle{Noise Stability Regularization for Improving {BERT} Fine-tuning}. In \bibinfo{booktitle}{\emph{Proceedings of the North American Chapter of the Association for Computational Linguistics: Human Language Technologies (NAACL-HLT)}}.
\newblock


\bibitem[\protect\citeauthoryear{Huang, Fang, et~al\mbox{.}}{Huang et~al\mbox{.}}{2024}]%
        {huang2024precision}
\bibfield{author}{\bibinfo{person}{Longwei Huang}, \bibinfo{person}{Chao Fang}, {et~al\mbox{.}}} \bibinfo{year}{2024}\natexlab{}.
\newblock \showarticletitle{A Precision-Scalable RISC-V DNN Processor with On-Device Learning Capability at the Extreme Edge}. In \bibinfo{booktitle}{\emph{Asia and South Pacific Design Automation Conference (ASP-DAC)}}. IEEE, \bibinfo{pages}{927--932}.
\newblock


\bibitem[\protect\citeauthoryear{Hubara, Courbariaux, et~al\mbox{.}}{Hubara et~al\mbox{.}}{2016}]%
        {BWN}
\bibfield{author}{\bibinfo{person}{Itay Hubara}, \bibinfo{person}{Matthieu Courbariaux}, {et~al\mbox{.}}} \bibinfo{year}{2016}\natexlab{}.
\newblock \showarticletitle{Binarized Neural Networks}. In \bibinfo{booktitle}{\emph{Advances in neural information processing systems (NeurIPS)}}, Vol.~\bibinfo{volume}{29}. \bibinfo{publisher}{Curran Associates, Inc.}
\newblock


\bibitem[\protect\citeauthoryear{Kim, Oh, et~al\mbox{.}}{Kim et~al\mbox{.}}{2024}]%
        {Minsik2024Layerwise-tcas1}
\bibfield{author}{\bibinfo{person}{Minsik Kim}, \bibinfo{person}{Kyoungseok Oh}, {et~al\mbox{.}}} \bibinfo{year}{2024}\natexlab{}.
\newblock \showarticletitle{A Low-Latency {FPGA} Accelerator for YOLOv3-Tiny With Flexible Layerwise Mapping and Dataflow}.
\newblock \bibinfo{journal}{\emph{{IEEE} Transactions on Circuits and Systems I: Regular Papers (TCAS-I)}} \bibinfo{volume}{71}, \bibinfo{number}{3} (\bibinfo{year}{2024}), \bibinfo{pages}{1158--1171}.
\newblock


\bibitem[\protect\citeauthoryear{Kim, Gholami, et~al\mbox{.}}{Kim et~al\mbox{.}}{2021}]%
        {I-BERT}
\bibfield{author}{\bibinfo{person}{Sehoon Kim}, \bibinfo{person}{Amir Gholami}, {et~al\mbox{.}}} \bibinfo{year}{2021}\natexlab{}.
\newblock \showarticletitle{{I-BERT:} Integer-only {BERT} Quantization}. In \bibinfo{booktitle}{\emph{Proceedings of International Conference on Machine Learning (ICML)}} \emph{(\bibinfo{series}{Proceedings of Machine Learning Research})}, Vol.~\bibinfo{volume}{139}. \bibinfo{publisher}{{PMLR}}, \bibinfo{pages}{5506--5518}.
\newblock


\bibitem[\protect\citeauthoryear{Li, Pandey, et~al\mbox{.}}{Li et~al\mbox{.}}{2020}]%
        {FTRANS}
\bibfield{author}{\bibinfo{person}{Bingbing Li}, \bibinfo{person}{Santosh Pandey}, {et~al\mbox{.}}} \bibinfo{year}{2020}\natexlab{}.
\newblock \showarticletitle{{FTRANS:} energy-efficient acceleration of {Transformers} using {FPGA}}. In \bibinfo{booktitle}{\emph{Proceedings of the ACM/IEEE International Symposium on Low Power Electronics and Design (ISLPED)}}. \bibinfo{publisher}{{ACM}}, \bibinfo{pages}{175--180}.
\newblock


\bibitem[\protect\citeauthoryear{Li and Gu}{Li and Gu}{2023}]%
        {I-ViT}
\bibfield{author}{\bibinfo{person}{Zhikai Li} {and} \bibinfo{person}{Qingyi Gu}.} \bibinfo{year}{2023}\natexlab{}.
\newblock \showarticletitle{I-ViT: Integer-only Quantization for Efficient Vision Transformer Inference}. In \bibinfo{booktitle}{\emph{{IEEE/CVF} International Conference on Computer Vision (ICCV)}}. \bibinfo{publisher}{{IEEE}}, \bibinfo{pages}{17019--17029}.
\newblock


\bibitem[\protect\citeauthoryear{Liu, Li, et~al\mbox{.}}{Liu et~al\mbox{.}}{2021}]%
        {FQ-BERT}
\bibfield{author}{\bibinfo{person}{Zejian Liu}, \bibinfo{person}{Gang Li}, {et~al\mbox{.}}} \bibinfo{year}{2021}\natexlab{}.
\newblock \showarticletitle{Hardware Acceleration of Fully Quantized {BERT} for Efficient Natural Language Processing}. In \bibinfo{booktitle}{\emph{Design, Automation \& Test in Europe Conference (DATE)}}. \bibinfo{publisher}{{IEEE}}, \bibinfo{pages}{513--516}.
\newblock


\bibitem[\protect\citeauthoryear{Liu, Oguz, et~al\mbox{.}}{Liu et~al\mbox{.}}{2022}]%
        {BiT}
\bibfield{author}{\bibinfo{person}{Zechun Liu}, \bibinfo{person}{Barlas Oguz}, {et~al\mbox{.}}} \bibinfo{year}{2022}\natexlab{}.
\newblock \showarticletitle{{BiT:} Robustly Binarized Multi-distilled Transformer}. In \bibinfo{booktitle}{\emph{Advances in neural information processing systems (NeurIPS)}}, Vol.~\bibinfo{volume}{35}. \bibinfo{pages}{14303--14316}.
\newblock


\bibitem[\protect\citeauthoryear{Liu, Wu, et~al\mbox{.}}{Liu et~al\mbox{.}}{2018}]%
        {Bi-Real}
\bibfield{author}{\bibinfo{person}{Zechun Liu}, \bibinfo{person}{Baoyuan Wu}, {et~al\mbox{.}}} \bibinfo{year}{2018}\natexlab{}.
\newblock \showarticletitle{Bi-Real Net: Enhancing the Performance of 1-Bit CNNs with Improved Representational Capability and Advanced Training Algorithm}. In \bibinfo{booktitle}{\emph{{European Conference on Computer Vision (ECCV)} {(15)}}}, Vol.~\bibinfo{volume}{11219}. \bibinfo{publisher}{Springer}, \bibinfo{pages}{747--763}.
\newblock


\bibitem[\protect\citeauthoryear{Lu, Wang, et~al\mbox{.}}{Lu et~al\mbox{.}}{2020}]%
        {Lu}
\bibfield{author}{\bibinfo{person}{Siyuan Lu}, \bibinfo{person}{Meiqi Wang}, {et~al\mbox{.}}} \bibinfo{year}{2020}\natexlab{}.
\newblock \showarticletitle{Hardware Accelerator for Multi-Head Attention and Position-Wise Feed-Forward in the {Transformer}}. In \bibinfo{booktitle}{\emph{IEEE International System-on-Chip Conference (SOCC)}}. \bibinfo{publisher}{{IEEE}}, \bibinfo{pages}{84--89}.
\newblock


\bibitem[\protect\citeauthoryear{Nguyen, Kim, et~al\mbox{.}}{Nguyen et~al\mbox{.}}{2021}]%
        {Duy2021Layer-Specific}
\bibfield{author}{\bibinfo{person}{Duy~Thanh Nguyen}, \bibinfo{person}{Hyun Kim}, {et~al\mbox{.}}} \bibinfo{year}{2021}\natexlab{}.
\newblock \showarticletitle{Layer-Specific Optimization for Mixed Data Flow With Mixed Precision in {FPGA} Design for CNN-Based Object Detectors}.
\newblock \bibinfo{journal}{\emph{{IEEE} Transactions on Circuits and Systems for Video Technology (TCSVT).}} \bibinfo{volume}{31}, \bibinfo{number}{6} (\bibinfo{year}{2021}), \bibinfo{pages}{2450--2464}.
\newblock


\bibitem[\protect\citeauthoryear{Park, Yoon, et~al\mbox{.}}{Park et~al\mbox{.}}{2020}]%
        {OPTIMUS}
\bibfield{author}{\bibinfo{person}{Junki Park}, \bibinfo{person}{Hyunsung Yoon}, {et~al\mbox{.}}} \bibinfo{year}{2020}\natexlab{}.
\newblock \showarticletitle{{OPTIMUS:} OPTImized matrix MUltiplication Structure for Transformer neural network accelerator}. In \bibinfo{booktitle}{\emph{Proceedings of Machine Learning and Systems (MLSys)}}. \bibinfo{publisher}{mlsys.org}.
\newblock


\bibitem[\protect\citeauthoryear{Qin, Ding, et~al\mbox{.}}{Qin et~al\mbox{.}}{2022}]%
        {BiBERT}
\bibfield{author}{\bibinfo{person}{Haotong Qin}, \bibinfo{person}{Yifu Ding}, {et~al\mbox{.}}} \bibinfo{year}{2022}\natexlab{}.
\newblock \showarticletitle{{BiBERT:} Accurate Fully Binarized {BERT}}. In \bibinfo{booktitle}{\emph{International Conference on Learning Representations (ICLR)}}.
\newblock


\bibitem[\protect\citeauthoryear{Qiu, Wang, et~al\mbox{.}}{Qiu et~al\mbox{.}}{2016}]%
        {jiantao2016Embedded-isfpga}
\bibfield{author}{\bibinfo{person}{Jiantao Qiu}, \bibinfo{person}{Jie Wang}, {et~al\mbox{.}}} \bibinfo{year}{2016}\natexlab{}.
\newblock \showarticletitle{Going Deeper with Embedded {FPGA} Platform for Convolutional Neural Network}. In \bibinfo{booktitle}{\emph{International Symposium on Field-Programmable Gate Arrays (ISFPGA)}}. \bibinfo{publisher}{{ACM}}, \bibinfo{pages}{26--35}.
\newblock


\bibitem[\protect\citeauthoryear{Rastegari, Ordonez, et~al\mbox{.}}{Rastegari et~al\mbox{.}}{2016}]%
        {XNOR-Net}
\bibfield{author}{\bibinfo{person}{Mohammad Rastegari}, \bibinfo{person}{Vicente Ordonez}, {et~al\mbox{.}}} \bibinfo{year}{2016}\natexlab{}.
\newblock \showarticletitle{XNOR-Net: ImageNet Classification Using Binary Convolutional Neural Networks}. In \bibinfo{booktitle}{\emph{{European Conference on Computer Vision (ECCV)} {(4)}}}, Vol.~\bibinfo{volume}{9908}. \bibinfo{publisher}{Springer}, \bibinfo{pages}{525--542}.
\newblock


\bibitem[\protect\citeauthoryear{Shao, Shi, et~al\mbox{.}}{Shao et~al\mbox{.}}{2024}]%
        {haikuo2024efficientViT}
\bibfield{author}{\bibinfo{person}{Haikuo Shao}, \bibinfo{person}{Huihong Shi}, {et~al\mbox{.}}} \bibinfo{year}{2024}\natexlab{}.
\newblock \showarticletitle{An FPGA-Based Reconfigurable Accelerator for Convolution-Transformer Hybrid EfficientViT}.
\newblock \bibinfo{journal}{\emph{CoRR}}  \bibinfo{volume}{abs/2403.20230} (\bibinfo{year}{2024}).
\newblock


\bibitem[\protect\citeauthoryear{Shen, Dong, et~al\mbox{.}}{Shen et~al\mbox{.}}{2020}]%
        {QBERT}
\bibfield{author}{\bibinfo{person}{Sheng Shen}, \bibinfo{person}{Zhen Dong}, {et~al\mbox{.}}} \bibinfo{year}{2020}\natexlab{}.
\newblock \showarticletitle{{Q-BERT:} Hessian Based Ultra Low Precision Quantization of {BERT}}. In \bibinfo{booktitle}{\emph{Proceedings of AAAI Conference on Artificial Intelligence (AAAI)}}. \bibinfo{publisher}{{AAAI} Press}, \bibinfo{pages}{8815--8821}.
\newblock


\bibitem[\protect\citeauthoryear{Song, Wang, et~al\mbox{.}}{Song et~al\mbox{.}}{2022}]%
        {UCViT}
\bibfield{author}{\bibinfo{person}{HongRui Song}, \bibinfo{person}{Ya Wang}, {et~al\mbox{.}}} \bibinfo{year}{2022}\natexlab{}.
\newblock \showarticletitle{UCViT: Hardware-Friendly Vision Transformer via Unified Compression}. In \bibinfo{booktitle}{\emph{IEEE International Symposium on Circuits and Systems (ISCAS)}}. \bibinfo{publisher}{{IEEE}}, \bibinfo{pages}{2022--2026}.
\newblock


\bibitem[\protect\citeauthoryear{Tian, Fang, et~al\mbox{.}}{Tian et~al\mbox{.}}{2023}]%
        {BEBERT}
\bibfield{author}{\bibinfo{person}{Jiayi Tian}, \bibinfo{person}{Chao Fang}, {et~al\mbox{.}}} \bibinfo{year}{2023}\natexlab{}.
\newblock \showarticletitle{{BEBERT}: Efficient and robust binary ensemble {BERT}}. In \bibinfo{booktitle}{\emph{IEEE International Conference on Acoustics, Speech and Signal Processing (ICASSP)}}. \bibinfo{publisher}{IEEE}, \bibinfo{pages}{1--5}.
\newblock


\bibitem[\protect\citeauthoryear{Tu, Wu, et~al\mbox{.}}{Tu et~al\mbox{.}}{2020}]%
        {tu2020evolver}
\bibfield{author}{\bibinfo{person}{Fengbin Tu}, \bibinfo{person}{Weiwei Wu}, {et~al\mbox{.}}} \bibinfo{year}{2020}\natexlab{}.
\newblock \showarticletitle{Evolver: A deep learning processor with on-device quantization--voltage--frequency tuning}.
\newblock \bibinfo{journal}{\emph{IEEE Journal of Solid-State Circuits (JSSC)}} \bibinfo{volume}{56}, \bibinfo{number}{2} (\bibinfo{year}{2020}), \bibinfo{pages}{658--673}.
\newblock


\bibitem[\protect\citeauthoryear{Vo, Le, et~al\mbox{.}}{Vo et~al\mbox{.}}{2022}]%
        {quang2022streaming-access}
\bibfield{author}{\bibinfo{person}{Quang~Hieu Vo}, \bibinfo{person}{Linh~Ngoc Le}, {et~al\mbox{.}}} \bibinfo{year}{2022}\natexlab{}.
\newblock \showarticletitle{A Deep Learning Accelerator Based on a Streaming Architecture for Binary Neural Networks}.
\newblock \bibinfo{journal}{\emph{{IEEE} Access}}  \bibinfo{volume}{10} (\bibinfo{year}{2022}), \bibinfo{pages}{21141--21159}.
\newblock


\bibitem[\protect\citeauthoryear{Wang, Singh, et~al\mbox{.}}{Wang et~al\mbox{.}}{2019}]%
        {GLUE}
\bibfield{author}{\bibinfo{person}{Alex Wang}, \bibinfo{person}{Amanpreet Singh}, {et~al\mbox{.}}} \bibinfo{year}{2019}\natexlab{}.
\newblock \showarticletitle{{GLUE:} {A} Multi-Task Benchmark and Analysis Platform for Natural Language Understanding}. In \bibinfo{booktitle}{\emph{International Conference on Learning Representations (ICLR)}}.
\newblock


\bibitem[\protect\citeauthoryear{Wang, Gong, et~al\mbox{.}}{Wang et~al\mbox{.}}{2022}]%
        {ViA}
\bibfield{author}{\bibinfo{person}{Teng Wang}, \bibinfo{person}{Lei Gong}, {et~al\mbox{.}}} \bibinfo{year}{2022}\natexlab{}.
\newblock \showarticletitle{{ViA:} {A} Novel Vision-{Transformer} Accelerator Based on {FPGA}}.
\newblock \bibinfo{journal}{\emph{{IEEE} Trans. Comput. Aided Des. Integr. Circuits Syst. (TCAD)}} \bibinfo{volume}{41}, \bibinfo{number}{11} (\bibinfo{year}{2022}), \bibinfo{pages}{4088--4099}.
\newblock


\bibitem[\protect\citeauthoryear{Wu, Liang, et~al\mbox{.}}{Wu et~al\mbox{.}}{2023}]%
        {xiaowu2023nonlinear}
\bibfield{author}{\bibinfo{person}{Xiao Wu}, \bibinfo{person}{Shuang Liang}, {et~al\mbox{.}}} \bibinfo{year}{2023}\natexlab{}.
\newblock \showarticletitle{ReAFM: A Reconfigurable Nonlinear Activation Function Module for Neural Networks}.
\newblock \bibinfo{journal}{\emph{IEEE Transactions on Circuits and Systems II: Express Briefs}} \bibinfo{volume}{70}, \bibinfo{number}{7} (\bibinfo{year}{2023}), \bibinfo{pages}{2660--2664}.
\newblock


\bibitem[\protect\citeauthoryear{Xu, Han, et~al\mbox{.}}{Xu et~al\mbox{.}}{2024}]%
        {OneBit}
\bibfield{author}{\bibinfo{person}{Yuzhuang Xu}, \bibinfo{person}{Xu Han}, {et~al\mbox{.}}} \bibinfo{year}{2024}\natexlab{}.
\newblock \showarticletitle{OneBit: Towards Extremely Low-bit Large Language Models}.
\newblock \bibinfo{journal}{\emph{CoRR}}  \bibinfo{volume}{abs/2402.11295} (\bibinfo{year}{2024}).
\newblock


\bibitem[\protect\citeauthoryear{Yang and Su}{Yang and Su}{2022}]%
        {EFA-Trans}
\bibfield{author}{\bibinfo{person}{Xin Yang} {and} \bibinfo{person}{Tao Su}.} \bibinfo{year}{2022}\natexlab{}.
\newblock \showarticletitle{{EFA-Trans:} An Efficient and Flexible Acceleration Architecture for {Transformers}}.
\newblock \bibinfo{journal}{\emph{Electronics}} \bibinfo{volume}{11}, \bibinfo{number}{21} (\bibinfo{year}{2022}).
\newblock


\bibitem[\protect\citeauthoryear{Zafrir, Boudoukh, et~al\mbox{.}}{Zafrir et~al\mbox{.}}{2019}]%
        {zafrir2019q8bert}
\bibfield{author}{\bibinfo{person}{Ofir Zafrir}, \bibinfo{person}{Guy Boudoukh}, {et~al\mbox{.}}} \bibinfo{year}{2019}\natexlab{}.
\newblock \showarticletitle{Q8bert: Quantized 8bit bert}. In \bibinfo{booktitle}{\emph{Fifth Workshop on Energy Efficient Machine Learning and Cognitive Computing-NeurIPS Edition (EMC2-NIPS)}}. IEEE, \bibinfo{pages}{36--39}.
\newblock


\bibitem[\protect\citeauthoryear{Zhang, Li, et~al\mbox{.}}{Zhang et~al\mbox{.}}{2019}]%
        {zhang2019compression-electronics}
\bibfield{author}{\bibinfo{person}{Min Zhang}, \bibinfo{person}{Linpeng Li}, {et~al\mbox{.}}} \bibinfo{year}{2019}\natexlab{}.
\newblock \showarticletitle{Optimized Compression for Implementing Convolutional Neural Networks on FPGA}.
\newblock \bibinfo{journal}{\emph{Electronics}} \bibinfo{volume}{8}, \bibinfo{number}{3} (\bibinfo{year}{2019}).
\newblock


\bibitem[\protect\citeauthoryear{Zhang, Roller, et~al\mbox{.}}{Zhang et~al\mbox{.}}{2022}]%
        {OPT}
\bibfield{author}{\bibinfo{person}{Susan Zhang}, \bibinfo{person}{Stephen Roller}, {et~al\mbox{.}}} \bibinfo{year}{2022}\natexlab{}.
\newblock \showarticletitle{{OPT:} Open Pre-trained Transformer Language Models}.
\newblock \bibinfo{journal}{\emph{CoRR}}  \bibinfo{volume}{abs/2205.01068} (\bibinfo{year}{2022}).
\newblock


\bibitem[\protect\citeauthoryear{Zhang, Hou, et~al\mbox{.}}{Zhang et~al\mbox{.}}{2020}]%
        {TernaryBERT}
\bibfield{author}{\bibinfo{person}{Wei Zhang}, \bibinfo{person}{Lu Hou}, {et~al\mbox{.}}} \bibinfo{year}{2020}\natexlab{}.
\newblock \showarticletitle{{TernaryBERT:} Distillation-aware Ultra-low Bit {BERT}}. In \bibinfo{booktitle}{\emph{Proceedings of the 2020 Conference on Empirical Methods in Natural Language Processing (EMNLP)}}. \bibinfo{publisher}{ACL}, \bibinfo{pages}{509--521}.
\newblock


\end{thebibliography}

\end{document}